\documentclass[11pt]{article}

\usepackage{amsmath, amsfonts, amsthm, amssymb, hyperref}
\usepackage{mathabx}
\usepackage{authblk, color, bm, graphicx}
\usepackage[margin = 1.5cm, font = small, labelfont = bf, labelsep = period]{caption}
\usepackage{xspace}
\usepackage[utf8]{inputenc}
\usepackage[round]{natbib}
\usepackage{rotating, multirow}
\usepackage{xcolor}
\usepackage{fullpage}[2cm]
\usepackage{mathtools}
\usepackage{paralist}
\usepackage{subfigure}
\RequirePackage{algorithm}
\RequirePackage{algorithmic}
\usepackage[textsize=tiny]{todonotes}
\usepackage[capitalize,noabbrev]{cleveref}
\usepackage{microtype}
\usepackage{subfigure}
\usepackage{comment}
\usepackage{appendix}

\theoremstyle{plain}
\newtheorem{theorem}{Theorem}[section]
\newtheorem{proposition}[theorem]{Proposition}

\newtheorem{corollary}[theorem]{Corollary}
\newtheorem{example}[theorem]{Example} % new add
\theoremstyle{definition}

\theoremstyle{remark}

\usepackage{enumitem}

\usepackage[T1]{fontenc}    % use 8-bit T1 fonts
\usepackage{hyperref}       % hyperlinks
\usepackage{url}            % simple URL typesetting
\usepackage{booktabs}       % professional-quality tables
\usepackage{nicefrac}       % compact symbols for 1/2, etc.
\usepackage{microtype}      % microtypography
\usepackage{xcolor}         % colors

\title{Distributionally Robust Federated Learning: An ADMM Algorithm}

\date{}

\author[1]{Wen Bai}
\author[1]{Yi Wong}
\author[1]{Xiao Qiao}
\author[1]{Chin Pang Ho}

\affil[1]{\small \textit{Department of Data Science, City University of Hong Kong, Hong Kong} \protect\\ \texttt{wen.bai@my.cityu.edu.hk}, \texttt{ywong692-c@my.cityu.edu.hk}, \protect\\ \texttt{xiaoqiao@cityu.edu.hk}, \texttt{clint.ho@cityu.edu.hk}}

\begin{document}

\maketitle

\begin{abstract}
Federated learning (FL) aims to train machine learning (ML) models collaboratively using decentralized data, bypassing the need for centralized data aggregation. Standard FL models often assume that all data come from the same unknown distribution. However, in practical situations, decentralized data frequently exhibit heterogeneity. We propose a novel FL model, Distributionally Robust Federated Learning (DRFL), that applies distributionally robust optimization to overcome the challenges posed by data heterogeneity and distributional ambiguity. We derive a tractable reformulation for DRFL and develop a novel solution method based on the alternating direction method of multipliers (ADMM) algorithm to solve this problem. Our experimental results demonstrate that DRFL outperforms standard FL models under data heterogeneity and ambiguity.
\end{abstract}

\section{Introduction} \label{introduction}

Data collection and data processing are increasingly becoming more distributed across multiple devices, organizations, and locations. Federated learning is a technique that enables collaborative model training with decentralized data. Model training occurs locally on distributed devices, or clients, while model updates are aggregated by the central server and shared across clients. For example, model parameters may be averaged across participating clients \citep{DBLP:journals/corr/McMahanMRA16}, such that all data are used to train the model without direct data sharing cross clients, preserving data privacy, data security, and data access restrictions. FL has emerged as an important field of research with impact in areas such as mobile devices \citep{lim2020federated}, healthcare \citep{joshi2022federated, nishio2019client}, and the internet of things \citep{nguyen2021federated}. 

Federated learning also has its challenges. Statistical heterogeneity, the situation in which data distribution across clients differs, poses a difficulty for federated learning methods that assume an i.i.d. data generating process for all clients. Such heterogeneity can lead to incorrect model prediction or embed certain biases if it is not properly accounted for. Federated learning also faces the issue of ambiguity, where the underlying data distributions across different clients are unknown. As a result, standard FL models often suffer from the issue of overfitting with poor out-of-sample performance, especially when the data are insufficient, noisy, or contaminated \citep{reisizadeh2020robust}.

Researchers have explored the applications of (distributionally) robust optimization for federated learning in recent years. \cite{mohri2019agnostic} introduce the concept of agnostic loss which calculates the worst-case weighted client loss. The authors propose a framework called agnostic federated learning with the aim of optimizing the worst-case weighted average performance across all clients. Building on this idea, \cite{deng2020distributionally} propose the distributionally robust federated averaging (DRFA) algorithm, an efficient method for handling the subpopulation shift. However, DRFA only takes into account the empirical distribution, which may differ significantly from the true distribution.
 
\cite{nguyen2022generalization} propose Wasserstein distributionally robust federated learning (WAFL) to address the issue of distributional ambiguity. They employ an ``aggregate-then-robustify'' approach, whereby they initially estimate a single probability distribution that represents all clients. An ambiguity set is then introduced to mitigate the impact induced by the estimation errors of this probability distribution. In this case, while a distributionally robust FL model is formulated, the associated optimization problem is not well-suited for designing distributed algorithms. Instead, an approximation algorithm is used, which is equivalent to solving a non-robust FL model with a modified loss function. Moreover, since this approach considers an aggregated/mixed probability distribution, it does not explicitly tackle the issue of heterogeneity. A simple example will be provided to further demonstrate the difference between WAFL and our proposed model.

To address the challenges of statistical heterogeneity and ambiguity simultaneously, we introduce a novel framework called \emph{Distributionally Robust Federated Learning (DRFL)}. In contrast to previous approaches, DRFL constructs an individual Wasserstein ambiguity set for each client, which enables each client to have its own, potentially unique data generating distribution. This design allows arbitrary levels of statistical heterogeneity among clients, setting DRFL apart from existing methods in federated learning.

A key contribution of our work is the derivation of a tractable reformulation of DRFL that facilitates further analysis. The resulting optimization problem differs structurally from the standard unconstrained empirical risk minimization problem with a finite sum of functions. This distinction prevents the efficient application of stochastic gradient descent (SGD) and its variants, as they may lead to high computational costs. As an alternative, we develop a novel solution method based on the alternating direction method of multipliers (ADMM) algorithm, and propose a splitting strategy to solve DRFL. The proposed algorithm has a natural interpretation for the client updates: In each iteration, every client updates its (regularized) worst-case expected loss, while the server updates the global parameter based on the information received from all clients.

We summarize the main contributions of our work as follows:

\begin{compactitem}
\item We propose a novel framework in federated learning, DRFL, capable of handling data heterogeneity and distributional ambiguity issues. We then derive a tractable reformulation of DRFL, making it suitable for further theoretical and algorithmic analysis.

\item We develop a new ADMM-based algorithm suitable for DRFL. Our approach ensures convergence to the optimal solution in the convex setting and to critical points in the non-convex setting, making it applicable to a wide range of learning problems.

\item We compare the empirical performance of DRFL with its non-robust counterpart (the nominal model) and other robust optimization methods. Our results demonstrate that DRFL outperforms other models on various problems.%tasks, highlighting its effectiveness in handling both statistical heterogeneity and uncertainty.
\end{compactitem}

The rest of this paper is organized as follows. Section~\ref{sec:related_work} reviews the related work, and Section~\ref{sec:prelim} provides background on federated learning. In Section~\ref{sec:DRFL}, we introduce Distributionally Robust Federated Learning and derive a tractable reformulation. In Section~\ref{sec:admm}, a customized ADMM algorithm is proposed to exactly solve DRFL. The numerical experiments in Section~\ref{sec:numerical} demonstrates the superior performance of the proposed model. 

\textbf{Notation.} We adopt boldface lowercase and uppercase letters for vectors and matrices, respectively. Special vectors include $\bm{0}$, which represents the vector of all zeros, and $\mathbf{e}$, which indicates the vector of all ones. The matrix $\bm{I}$ denotes the unit matrix. The inner product of vectors $\bm{w}$ and $\bm{x}$ is denoted by $\left\langle \bm{w}, \bm{x} \right\rangle$. We use the tilde sign to represent random variables. The dual norm of any given norm $\Vert \cdot \Vert$ is denoted by $\Vert \cdot \Vert_*$. The notation $\Delta_S = \{\bm{q}\in\mathbb{R}^S_+ ~|~ \mathbf{e}^\top \bm{q} = 1 \}$ is used to represent probability simplex. The operator $[\cdot ]_+$ represents $\max \{ \cdot ,0 \}$.

\section{Related Work} \label{sec:related_work}

Federated learning (FL) was initially introduced by \cite{DBLP:journals/corr/McMahanMRA16}. They proposed the Federated Averaging (FedAvg) algorithm, which utilizes SGD in parallel on a randomly sampled subset of clients at every iteration. Then the server updates the model parameters by averaging the updated parameters from clients, and so it allows for the training of models on decentralized data while preserving privacy.

In general, the challenges faced by FL can be broadly categorized into two areas: system challenges and statistical heterogeneity. Several prior works have proposed various approaches to tackle system challenges, with a primary focus on improving communication efficiency in federated learning \citep{konevcny2015federated, konevcny2016federated, li2020federated, reisizadeh2020fedpaq, smith2017federated, suresh2017distributed}. In terms of statistical heterogeneity, FedAvg generalizes poorly on the unseen distribution of new clients and non-i.i.d. data \citep{zhao2018federated}. Several studies offered improvements to the FedAvg \citep{li2019convergence, sattler2019robust, zhao2018federated}, especially on enhancing either the global aggregation step \citep{chen2020fedbe, lin2020ensemble, reddi2020adaptive, wang2020federated, yurochkin2019bayesian} or the local updates \citep{malinovskiy2020local, reisizadeh2020robust, wang2020tackling}. Many frameworks are proposed, such as FedProx \citep{li2020federated}, SCAFFOLD \citep{karimireddy2020scaffold}, FedNova \citep{wang2020tackling}, FedDyn \citep{acar2021federated}, MOON \citep{li2021model}, FedSkip \citep{fan2022fedskip}, and FedDC \citep{gao2022feddc}. Alternatively, \cite{smith2017federated} and \cite{corinzia2019variational} explore the application of multi-task learning to tackle statistical heterogeneity in FL settings. Additionally, several studies propose personalized federated learning approaches to enhance communication efficiency and penalize local models, such as \citep{chen2021bridging, collins2021exploiting, deng2020adaptive, hanzely2020lower, li2021ditto, mansour2020three}.

This paper utilizes distributionally robust optimization (DRO) to overcome the challenge of statistical heterogeneity. DRO optimizes the worst-case performance over an ambiguity set of distributions, providing reliable decisions under uncertainty and ambiguity \citep{goh2010distributionally}. In recent years, DRO has gained significant attention in machine learning \citep{namkoong2016stochastic}. In particular, Wasserstein DRO has gained attention in the areas of machine learning and data-driven optimization \citep{farokhi2022distributionally, shafieezadeh2019regularization, sinha2017certifying, smirnova2019distributionally, staib2019distributionally}, offering finite sample guarantees and accommodating unknown true distributions with larger support \citep{mohajerin2018data}.

Several seminal works have explored the application of (distributionally) robust optimization in the context of FL. \cite{mohri2019agnostic} introduce agnostic federated learning (AFL), minimizing the worst-case weighted sum of empirical loss among all clients. \cite{deng2020distributionally} develop a communication-efficient distributed algorithm for a specific case of AFL, where the uncertainty set of weights is defined as the probability simplex. \cite{nguyen2022generalization} propose a Wasserstein distributionally robust FL model for the mixed empirical distribution that represents the data generating distribution for all clients. 
\cite{reisizadeh2020robust} address affine distribution shifts and apply robust optimization to minimize loss under the worst perturbation. The proposed DRFL approach generalizes these methods, considering ambiguity at the individual client level without specific parametric perturbations. This flexibility enables comprehensive handling of uncertainties, accommodating diverse variability across clients. Moreover, the proposed framework does not make any assumption regarding whether training and testing distribution have the same support, so both subpopulation shift and domain shift are considered.

While this paper primarily addresses heterogeneity and ambiguity in standard FL, related works apply DRO to other contexts, such as robust federated meta-learning \citep{lin2020collaborative} and optimizing robustness under local differential privacy \citep{shi2022distributionally}.

\section{Preliminaries} \label{sec:prelim}
Consider a federated learning setting with a set of clients (also known as devices). We denote by $\mathcal{S} = \{1,2,\dots,S\}$ the set of clients, in which each client~$s$ has it own dataset $\{ (\hat{\bm{x}}_{i},\hat{y}_{i}) \}_{i=1}^{N_s}$ with $N_s$ training samples, for all $s\in\mathcal{S}$. Here, $\hat{\bm{x}}_{si}$ and $\hat{y}_{si}$ are the feature vector and label of the $i$th sample in client $s$, respectively. We denote by $\Xi \subseteq \mathbb{R}^n \times \mathbb{R}$ the support set of the samples, and thus $(\hat{\bm{x}}_{i},\hat{y}_{i}) \in \Xi$. Given a loss function $L$ and a function class $\mathcal{F} = \{f_{\bm{w}}:\bm{w}\in\mathcal{W} \subseteq \mathbb{R}^m \}$, the standard federated learning problem is to seek for the model parameters $\bm{w}$ by minimizing the following empirical risk minimization (ERM) problem
\begin{equation} \label{eq:non_robust_problem}
\min_{\bm{w} \in \mathcal{W}}\; \frac{1}{S} \sum_{s\in\mathcal{S}} \frac{1}{N_s} \sum_{i \in \mathcal{I}_s} L( f_{\bm{w}}(\hat{\bm{x}}_{si}), \hat{{y}}_{si}),
\end{equation}
where $\mathcal{I}_s = \{1,2\dots , N_s \}$. The above model covers a wide range of supervised learning problems; for example, consider regression problem with linear model where $f_{\bm{w}} (\bm{x}) = \left \langle \bm{w} , \bm{x} \right \rangle$ and $L( f_{\bm{w}}(\bm{x}), y) = L(\left \langle \bm{w} , \bm{x} \right \rangle - y)$, problem \eqref{eq:non_robust_problem} covers the following special cases.
\begin{itemize} %[leftmargin=*]
\item {Huber Regression (HR):} 
\begin{equation*}
L(z) = \left\{
\begin{array}{ll}
\displaystyle z^2 / 2 & \text{if } \vert z \vert \leq \epsilon \\
\displaystyle \epsilon (\vert z \vert - \epsilon / 2) & \text{otherwise,}
\end{array}
\right.
\end{equation*}
where $\epsilon > 0$ is a user-specified threshold parameter.  
\item {Support Vector Regression (SVR):} %$L_{\rm R}(z) = \max \{ 0, \vert z \vert - \epsilon \},$
\begin{equation*}
L(z) = \max \{ 0, \vert z \vert - \epsilon \},
\end{equation*}
with tolerance level $\epsilon \geq 0$.
\item {Quantile Regression:} %$L_{\rm R}(z) = \max \{ -\epsilon z, (1-\epsilon) z \},$
\begin{equation*}
L(z) = \max \{ -\epsilon z, (1-\epsilon) z \},
\end{equation*}
where $\epsilon \in [0, 1]$ is the quantile parameter.
\end{itemize}

For classification problem with linear model where $f_{\bm{w}} (\bm{x}) = \left \langle \bm{w} , \bm{x} \right \rangle$ and $L( f_{\bm{w}}(\bm{x}), y) = L(y \left \langle \bm{w}, \bm{x} \right \rangle)$, problem \eqref{eq:non_robust_problem} covers the following special cases.

\begin{itemize} %[leftmargin=*]
\item {Support Vector Machine (SVM):}% $L(z) = \max \{ 0, 1 - z \}.$
\begin{equation*}
L(z) = \max \{ 0, 1 - z \}.
\end{equation*}
\item {SVM with Smooth Hinge Loss:} 
\begin{equation*}
L(z) = \left\{
\begin{array}{ll}
\displaystyle 1/2 - z & \text{if } z \leq 0 \\
\displaystyle (1-z)^2/2 & \text{if } 0 < z < 1 \\
\displaystyle 0 & \text{otherwise.}
\end{array}
\right.
\end{equation*}
\item {Logistic Regression:} %$L_{\rm C}(z) = \log(1 + e^{-z})$.
\begin{equation*}
L(z) = \log(1 + e^{-z}).
\end{equation*}
\end{itemize}

%, such as Huber regression (HR), and support vector machine (SVM) (see Appendix~\ref{appendix:loss_function_def}).

In the setting of federated learning, training samples are stored separately across different clients, and each individual dataset $\{ (\hat{\bm{x}}_{i},\hat{y}_{i}) \}_{i=1}^{N_s}$ in client $s$ is not shareable to other clients $s'$ during the training period. Fortunately, despite being a generic model, problem~\eqref{eq:non_robust_problem} exhibits strong mathematical structure and is a sum of functions where each function corresponds to a training sample; thus, by leveraging the idea of SGD, FedAvg was proposed to solve \eqref{eq:non_robust_problem} in a federated setting \citep{DBLP:journals/corr/McMahanMRA16} where each client is responsible to the gradient updates that are associated with its training samples.

\section{Distributionally Robust Federated Learning} \label{sec:DRFL}

While problem~\eqref{eq:non_robust_problem} serves as a generalization of standard supervised learning with the consideration that data is stored separately across different clients, it suffers the from the same issue of overfitting as other supervised learning models, due to the fact that the data generating distribution is unknown. In fact, the ERM formulation in \eqref{eq:non_robust_problem} is only an approximation of the stochastic optimization problem $\inf_{\bm{w} \in \mathcal{W}}\; \mathbb{E}^\mathbb{P} [ L( f_{\bm{w}}(\tilde{\bm{x}}), \tilde{y})]$, where $\mathbb{P}$ is the underlying unknown probability distribution that generates the training samples $\{ (\hat{\bm{x}}_{i},\hat{y}_{i}) \}_{i \in \mathcal{I}_s}$, for all $s\in\mathcal{S}$. 

In addition to the aforementioned issue of ambiguity, applications of federated learning often face another challenge of data heterogeneity. Due to the decentralized nature of the data, training samples of each individual dataset are often collected by each client, and they may not be identically distributed compared to training samples of other clients; that is, for each client $s\in\mathcal{S}$, the samples $\{ (\hat{\bm{x}}_{i},\hat{y}_{i}) \}_{i \in \mathcal{I}_s}$ are generated via some distribution $\mathbb{P}_s$ which is not necessary same as $\mathbb{P}_{s'}$ if $s'\neq s$.

\subsection{Model}

To overcome these challenges, we propose the Distributionally Robust Federated Learning (DRFL)
\begin{equation} \label{eq:origin_problem}
\inf_{\bm{w} \in \mathcal{W}}\; \sup_{\mathbb{P} \in \mathcal{P}}\; \mathbb{E}^\mathbb{P} [ L( f_{\bm{w}}(\tilde{\bm{x}}), \tilde{y}) ],
\end{equation}
and we consider the ambiguity set $\mathcal{P}$ to be
\begin{equation*} %\label{eq:ambiguity_set}
\mathcal{P} = \left\{
\displaystyle \mathbb{P} \in \mathcal{P}^0(\Xi \times \mathcal{S}) \left|
\begin{array}{ll}
((\tilde{\bm{x}},\tilde{y}),\tilde{s}) \in \mathbb{P}, \; \bm{q} \in \mathcal{Q} \\
\displaystyle \mathbb{P}(W(\mathbb{P}_s,\hat{\mathbb{P}}_s) \leq \rho_s \vert \tilde{s} = s) = 1 & \forall s \in \mathcal{S} \\
\displaystyle \mathbb{P}(\tilde{s} = s) = q_s & \forall s \in \mathcal{S}
\end{array}
\right.
\right\}, 
\end{equation*}
where $\mathcal{P}^0(\Xi \times \mathcal{S})$ is the set of all possible probability distributions supported on $\Xi \times \mathcal{S}$, $\mathcal{Q} \subseteq \mathbb{R}^S$ is the set of all possible weight parameters, and $W(\mathbb{P}_s, \hat{\mathbb{P}}_s)$ represents the Wasserstein distance of $\mathbb{P}_s$ from empirical distribution $\hat{\mathbb{P}}_s$; that is, $W(\mathbb{P},\hat{\mathbb{P}}) := \inf_{\Pi} \left \{ \int_{\Xi^2} d(\xi,\hat{\xi}) \; \Pi(d\xi,d\hat{\xi}) \right \}$. Here, $\Pi$ is a joint distribution of $\xi$ and $\hat{\xi}$ with marginals $\mathbb{P}$ and $\hat{\mathbb{P}}$, respectively. Notice $d$ is a metric on the support $\Xi$, and different applications (e.g., regression or classification) would have different metric $d$. We refer interested reader to Appendix~\ref{appendix:loss_funcs} for more details.

Problem \eqref{eq:origin_problem} has a natural interpretation in the setting of federated learning. In DRFL, one seeks for the model parameter $\bm{w}$ that optimizes the worst-case expected loss among all distributions in the ambiguity set $\mathcal{P}$, which contains distributions that are equal to the client's data generating distribution $\mathbb{P}_s$ with probability $q_s$. Both $\mathbb{P}_s$ and $\bm{q}$ are unknown; the former is assumed to reside within a Wasserstein ball with center $\hat{\mathbb{P}}_s$ and radius $\rho_s$, while the latter is assumed to reside within a prescribed set $\mathcal{Q}$. In particular, \eqref{eq:origin_problem} can be re-expressed as 
\begin{equation*}
\inf_{\bm{w} \in \mathcal{W}} \; \max_{q\in\mathcal{Q}} \; \sum_{s\in\mathcal{S}} q_s \cdot \sup_{\mathbb{P}_s \in \mathcal{P}_s}\; \mathbb{E}^{\mathbb{P}_s} [ L( f_{\bm{w}}(\tilde{\bm{x}}), \tilde{y}) ], \quad\text{where}\quad \mathcal{P}_s = \left\{ \mathbb{P}_s \in \mathcal{P}^0(\Xi) ~|~ W(\mathbb{P}_s,\hat{\mathbb{P}}_s) \leq \rho_s \right\}.
\end{equation*}
Therefore, the ambiguity set $\mathcal{P}$ explicitly model heterogeneity by allowing each $\mathbb{P}_s$ to be different from others and reside in its own ambiguity set $\mathcal{P}_s$. Moreover, as opposed to most existing FL methods that fix the weighting parameter $\bm{q}$, DRFL does not require the $\bm{q}$ to be explicitly specified, which is preferable when the data in each client is not generated at the same rate. 

DRFL provides a generalization to several existing frameworks. In particular, when $\rho_s = 0$ for all $s\in\mathcal{S}$, DRFL does not consider distributional ambiguity and recover the model proposed in \cite{mohri2019agnostic}. If one further specifies that $\mathcal{Q} = \Delta_S$, then DRFL is equivalent to the model studied in \cite{deng2020distributionally}. To recover the distributionally robust model introduced in \cite{nguyen2022generalization}, one could focus on distributions inside a Wasserstein ball which is centered at the aggregated empirical distribution $\hat{\mathbb{P}}_{\text{agg}} = \sum_{s\in\mathcal{S}} q_s \cdot \hat{\mathbb{P}}_s$ with radius $\rho$, rather than separately consider individual empirical distributions $\mathbb{P}_s$. To highlight the merits of our proposed DRFL compared to WAFL, consider the following simple numerical example, where the details of the numerical setup could be found in Appendix~\ref{appendix:example_problem_setting}.

\begin{example} \label{example:volume_compare}
\begin{figure}[!htb] 
\centering
\begin{minipage}{5.5cm}
\centering
\includegraphics[width=1\textwidth]{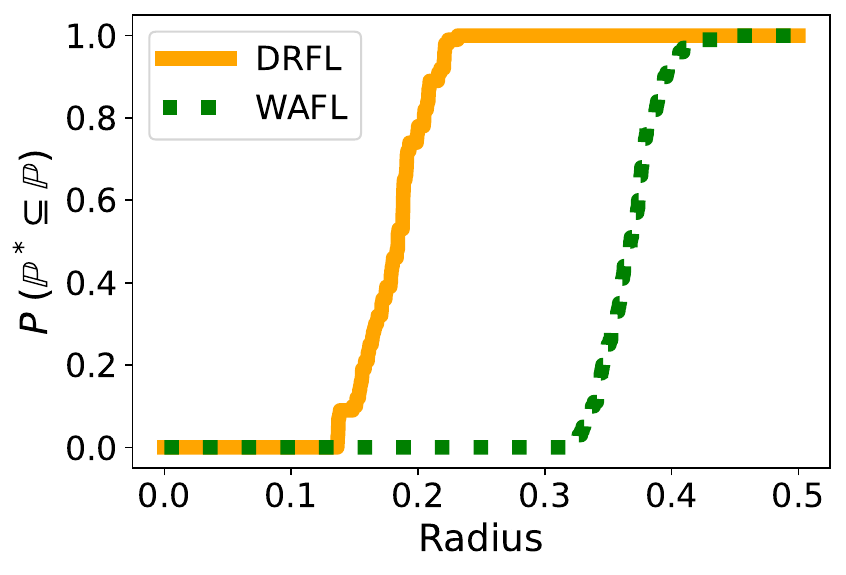}
\end{minipage}
\hspace{0.5cm}
\begin{minipage}{5.5cm}
\centering
\includegraphics[width=1\textwidth]{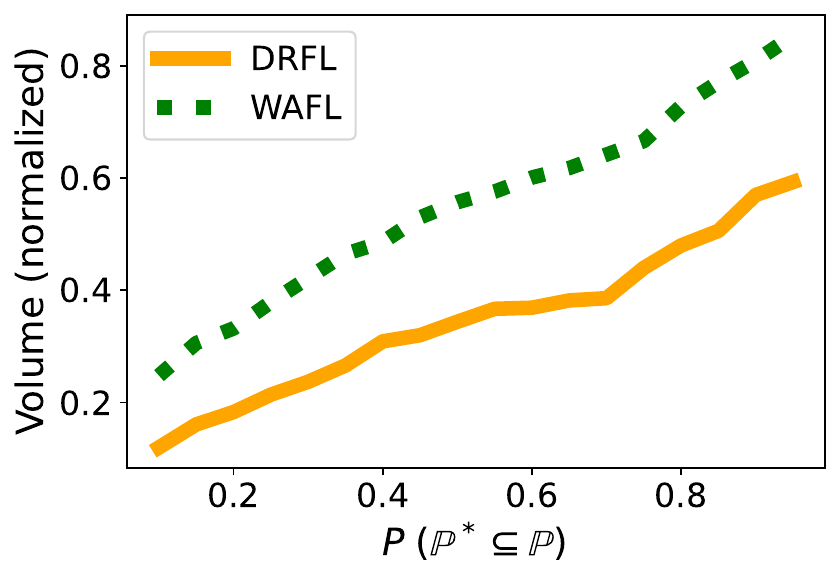}
\end{minipage}
\caption{$P (\mathbb{P}^\star \subseteq \mathbb{P})$ in different radius for two models (left) and the normalized distribution volume inside the ambiguity sets of two models in different guarantee levels to contain $\mathbb{P}^\star$ (right).}
\label{fig:example}
\end{figure}
Consider a simple system containing two clients with distributions $\mathbb{P}_1^\star$ and $\mathbb{P}_2^\star$ and their weights $q_1$ and $q_2$, respectively. By generating data from the underlying distribution $\mathbb{P}^\star = q_1 \cdot \mathbb{P}_1^\star + q_2 \cdot \mathbb{P}_2^\star$, we estimate empirical distributions $\hat{\mathbb{P}}_1$ and $\hat{\mathbb{P}}_2$ which could be used to construct the ambiguity sets $\mathcal{P}$ and $\mathcal{P}_W$ of DRFL and WAFL, respectively, where the center of $\mathcal{P}_W$ is $\hat{\mathbb{P}}_{agg} = \hat{q}_1 \cdot \hat{\mathbb{P}}_1 + \hat{q}_2 \cdot \hat{\mathbb{P}}_2$. Then, we first examine the probability of containing the $\mathbb{P}^\star$ for fixed radius in $\mathcal{P}$ and $\mathcal{P}_W$, where the result is shown in  Figure~\ref{fig:example} (Left). The result demonstrates that, WAFL always needs a larger radius when both models achieve the same performance guarantee. Figure~\ref{fig:example} (Right) further shows that, when both ambiguity sets $\mathcal{P}$ and $\mathcal{P}_W$ has the same confidence of containing the unknown true distribution $\mathbb{P}^\star$, the (normalized) volume of $\mathcal{P}_W$ is larger than $\mathcal{P}$.
\end{example}

Example~\ref{example:volume_compare} shows that, under the same statistical guarantee, WAFL tends to construct an ambiguity set $\mathcal{P}_W$ with higher volume compared to our proposed ambiguity set. This implies that it is more likely that WAFL would consider and optimize over the worst-case scenario that is unrealistic or highly unlikely to happen in practice; thus, providing more conservative solutions compared to DRFL.

\subsection{Reformulation} \label{sec:general_reform}

To solve problem \eqref{eq:origin_problem}, we apply techniques in Wasserstein distributionally robust optimization \citep{shafieezadeh2019regularization} and derive its tractable reformulation. To do so, we assume that for every fixed $\bm{w}$, the integrand $L( f_{\bm{w}}(\tilde{\bm{x}}), \tilde{y})$ in $\mathbb{E}^\mathbb{P} [ L( f_{\bm{w}}(\tilde{\bm{x}}), \tilde{y})]$ is bounded above by a Lipschitz continuous function; this assumption is mild, and all examples introduced in Section~\ref{sec:prelim} satisfy this assumption.

\begin{proposition} \label{prop:maxinf_omega}
For every $s\in\mathcal{S}$ and any fixed $\bm{w}\in\mathcal{W}$, 
\begin{equation*}
\sup_{\mathbb{P}_s \in \mathcal{P}_s}\mathbb{E}^{\mathbb{P}_s} [ L( f_{\bm{w}}(\tilde{\bm{x}}), \tilde{y}) ] = \inf_{\lambda_s, \bm{\alpha}_s} \Big \{ \rho_s \lambda_s + \frac{1}{N_s} \mathbf{e}^\top \bm{\alpha}_s: (\lambda_s, \bm{\alpha}_s, \bm{w}) \in \Omega_s \Big \},
\end{equation*}
where $\Omega_s \subseteq \mathbb{R}_+\times \mathbb{R}^{N_s}\times \mathcal{W}$ is defined as the following set
\begin{equation*}
\left\{ 
( \lambda_s, \; \bm{\alpha}_s, \; \bm{w}): \sup_{(\bm{x},y) \in \Xi }  L( f_{\bm{w}}(\bm{x}), y)  - \lambda_s d((\bm{x},y),(\hat{\bm{x}}_{si},\hat{y}_{si})) \leq \alpha_{si}, \; \forall i \in \mathcal{I}_s
\right\}.
\end{equation*}

In particular, problem~\eqref{eq:origin_problem} is equivalent to
\begin{equation} \label{eq:maxinf_omega}
\inf_{\bm{w}\in\mathcal{W}} \; \max_{\bm{q}\in\mathcal{Q}} \; \sum_{s\in\mathcal{S}} q_s \cdot \inf_{\lambda_s, \bm{\alpha}_s} \Big \{ \rho_s \lambda_s + \frac{1}{N_s} \mathbf{e}^\top \bm{\alpha}_s: \; (\lambda_s, \bm{\alpha}_s, \bm{w}) \in \Omega_s \Big \}.
\end{equation}
\end{proposition}

Notice that different applications (e.g., SVM or logistics regression) would lead of different $\Omega_s$, as it depends on both the loss function $L$ and the function $f_{\bm{w}}$. In many cases, including all the examples covered in Section~\ref{sec:prelim}, one could further simplify $\Omega_s$ based on the particular structure of $L$, $f_{\bm{w}}$, and $d$. We refer interested readers to Appendix~\ref{appendix:loss_funcs} for a detailed discussion on these simplifications. On the other hand, problem~\eqref{eq:maxinf_omega} offers a unified formulation for various applications. This is because all the problem-specific mathematical structure is now encapsulated in $\Omega_s$, while other parts of \eqref{eq:maxinf_omega} remain generic and adaptable to different cases.

The above proposition reformulates the problem \eqref{eq:non_robust_problem} into a min-max-min problem where the inner minimization problem is decomposed into $S$ smaller optimization problems. In particular, all the training samples $\{ (\hat{\bm{x}}_{i},\hat{y}_{i}) \}_{i=1}^{N_s}$ in client $s$ are only relevant to the $s$th inner minimization problem. Therefore, one could apply idea from algorithms for solving minimax problems, such as double-loop method \citep{nouiehed2019solving}, which allows one to train DRFL in a federated setting.

Using double-loop methods to solve \eqref{eq:maxinf_omega}, however, would lead to algorithms that have two layers of subproblems, which is not ideal from a computational perspective. In what follows, we take one step further to simplify the problems. Notice that for any fixed $\bm{w}\in\mathcal{W}$, $\Omega_s$ is convex in $(\lambda_s,\bm{\alpha}_s)$, even if $L$ or $f_{\bm{w}}$ are non-convex. Therefore, by duality and minimax theorem, we obtain the following.

\begin{theorem} \label{thm:main_reform}
Suppose $\mathcal{Q} = \{\bm{q}\in\Delta_S ~|~ \Vert \bm{q} - \hat{\bm{q}} \Vert_p \leq \theta \} $ for some $\hat{\bm{q}} \in \Delta_S$, $\theta > 0$, and $p \geq 1$. Then Problem~\eqref{eq:origin_problem} is equivalent to
\begin{equation} \label{eq:main_theorem}
\begin{array}{l@{\quad}ll}
\inf & \displaystyle \hat{\bm{q}}^\top (\bm{z} + \bm{\eta}) + \theta \cdot \Vert \bm{z} +\gamma  \mathbf{e} +\bm{\eta} \Vert_{p_*} & \\
\text{\rm s.t.} & \displaystyle z_s = \rho_s \lambda_s + \frac{1}{N_s} \mathbf{e}^\top \bm{\alpha}_s, \quad \ ( \lambda_s, \bm{\alpha}_s, \bm{w}) \in \Omega_s & \forall s\in\mathcal{S} \\
& \displaystyle \bm{w}\in \mathcal{W}, \; \gamma \in \mathbb{R}, \; \bm{\eta} \in \mathbb{R}^S_+, \; \bm{z} \in \mathbb{R}^S, \; \lambda_s \in \mathbb{R}, \; \bm{\alpha}_s \in \mathbb{R}^{N_s} & \forall s \in \mathcal{S},
\end{array}
\end{equation}
where $\Omega_s$ is defined as in Proposition~\ref{prop:maxinf_omega}.
\end{theorem}

In Theorem~\ref{thm:main_reform}, we specify $\mathcal{Q}$ to be an intersection of a probability simplex $\Delta_S$ and a $\ell_p$-norm ball with center $\hat{\bm{q}}$ and radius $\theta$. Two reasonable choices for setting $\hat{\bm{q}}$ are $\hat{q}_s = N_s / \sum_{s'\in\mathcal{S}} N_{s'}$ and $\hat{q}_s = 1 / S$ for all $s\in\mathcal{S}$, which correspond to assigning weights $q_s$ as a proportion to the total data in client $s$ and considering all clients to be equally important, respectively.

At the first glance, problem~\eqref{eq:main_theorem} may not exhibit a natural structure for federated learning, which requires clients do not share their data when optimizing the model parameter $\bm{w}$. As we will show in the next section, however, one could design a scalable first-order method to optimize problem~\eqref{eq:main_theorem} in a federated setting.

\section{An ADMM-Based Algorithm} \label{sec:admm}

In this section, we propose an efficient first-order method to solve our proposed DRFL based on the ADMM algorithm \citep{beck2017first}. In particular, we provide a novel splitting and apply the alternating direction linearized proximal method of multipliers (AD-LPMM) algorithm to solve~\eqref{eq:main_theorem}, where additional details are provided in Appendix~\ref{proof:add_detail_adlpmm}. Our method inherits the properties of AD-LPMM, and so it is exact and is guaranteed to converge. If \eqref{eq:main_theorem} is convex, our algorithm converges to optimality at an $\mathcal{O}(1/k)$ rate ~\cite{beck2017first}. More examples and details are provided in Appendix~\ref{appendix:loss_funcs}. If \eqref{eq:main_theorem} is non-convex, the algorithm nevertheless converges to critical points \citep{bolte2014proximal}.

We first clarify that SGD and its variants are not well-suited to solve DRFL exactly. Notice that \eqref{eq:main_theorem} is a constrained optimization problem where all samples are parameters of the constraints; thus, the projection step in any SGD-like method would require simultaneous access to all data, violating the nature of FL settings. Similarly, using off-the-shelf solvers (e.g. Gurobi) to solve \eqref{eq:main_theorem} is also not applicable in FL settings. Moreover, these solvers suffer from poor scalability.% with increasing data volume. 

As mentioned before, related algorithms such as double loop methods can be used to solve~\eqref{eq:maxinf_omega} (which is equivalent to \eqref{eq:main_theorem}), but they will lead to the explosion of sub-problems. In particular, double loop methods would treat \eqref{eq:maxinf_omega} as a single minimization problem where at each iteration, one has to solve a sub-problem which itself is a maximin problem, which can be solved using another double loop method. As we apply double loop method recursively, the number of iterations required should be quadratic to the number of iterations needed in usual first order methods.

To this end, we propose the AD-LPMM algorithm to solve \eqref{eq:main_theorem}; we first reformulate \eqref{eq:main_theorem} and obtain
\begin{equation} \label{eq:admm_split}
\begin{array}{l@{\quad}ll}
\min & \displaystyle \hat{\bm{q}}^\top\bm{t} + \theta \cdot \Vert \bm{t} \Vert_{p_*} - \gamma & \\
\text{\rm s.t.} & \bm{t} = \bm{z} +\gamma  \mathbf{e} +\bm{\eta}, \qquad \quad \displaystyle \bm{w} = \hat{\bm{w}}_s & \forall s\in\mathcal{S} \\
& \displaystyle z_s = \rho_s \lambda_s + \frac{1}{N_s} \mathbf{e}^\top \bm{\alpha}_{s}, \; ( \lambda_s, \bm{\alpha}_s, \hat{\bm{w}}_s) \in \Omega_s & \forall s\in\mathcal{S} \\
& \displaystyle \bm{w}\in \mathbb{R}^m, \; \bm{z} \in \mathbb{R}^S, \; \bm{t} \in \mathbb{R}^S, \; \gamma \in \mathbb{R} \\
& \displaystyle \bm{\eta} \in \mathbb{R}^S_+, \;  \lambda_s \in \mathbb{R},  \bm{\alpha}_s  \in \mathbb{R}^{N_s},  \hat{\bm{w}}_s \in \mathcal{W} & \forall s\in\mathcal{S},
\end{array}
\end{equation}
from which we can split primal variables into two groups, $(\bm{w},\bm{t},\bm{z})$ and $(\bm{\eta},\gamma,\{\lambda_s\},\{\bm{\alpha}_s\},\{\hat{\bm{w}}_s\})$, with separate primal updates. Since AD-LPMM is based on the optimization of the augmented Lagrangian, we assign $\bm{\sigma}$, $\{\bm{\psi}_s\}$, and $\{\bm{\zeta}_s\}$ to be the dual variables that are correspond to the linear equality constraints in problem~\eqref{eq:admm_split}. In what follows, we will describe the proposed algorithm that is summarized in Algorithm~\ref{alg:adlpmm}. We refer interested readers to the Appendix~\ref{proof:add_detail_adlpmm} for a detailed explanation of the mathematical details of this proposed algorithm.

\begin{algorithm}[!htb]
\caption{AD-LPMM for Problem~\eqref{eq:main_theorem}} \label{alg:adlpmm}
\begin{algorithmic}[1]
\STATE {\bfseries Input:} Set $k = 0$, stepsize $c > 0$, initial value $\bm{w}^0$, $\bm{t}^0$, $\bm{z}^0$, $\bm{\eta}^0$, $\gamma^0$, $\lambda_s^0$, $\bm{\alpha}_s^0$, $\hat{\bm{w}}_s^0$
\WHILE{Not satisfying stopping criterion} 

\STATE \texttt{// Primal update (run on server)}

\STATE $\bm{w}^{k+1} \leftarrow \mathfrak{P}_{\bm{w}}(\{\hat{\bm{w}}_s^k\}, \{\bm{\psi}_s^k\})$ \texttt{// see equation~\eqref{eq:operator_w}}

\STATE $\bm{z}^{k+1} \leftarrow \mathfrak{P}_{\bm{z}}(\bm{t}^k, \bm{z}^k, \bm{\eta}^k, \gamma^k, \{\lambda_s^k\}, \{\bm{\alpha}_s^k\}, \bm{\zeta}^k, \bm{\sigma}^k)$ \texttt{// see equation~\eqref{eq:operator_z}}

\STATE $\bm{t}^{k+1} \leftarrow \mathfrak{P}_{\bm{t}}(\bm{t}^k, \bm{z}^k, \bm{\eta}^k, \gamma^k, \bm{\sigma}^k)$ \texttt{// see equation~\eqref{eq:operator_t}}

\STATE $\bm{\eta}^{k+1} \leftarrow \mathfrak{P}_{\bm{\eta}}(\bm{t}^{k+1}, \bm{z}^{k+1}, \bm{\eta}^k, \gamma^k, \bm{\sigma}^k)$ \texttt{// see equation~\eqref{eq:operator_eta}}

\STATE $\gamma^{k+1} \leftarrow \mathfrak{P}_{\gamma}(\bm{t}^{k+1}, \bm{z}^{k+1}, \bm{\eta}^k, \gamma^k, \bm{\sigma}^k)$ \texttt{// see equation~\eqref{eq:operator_gamma}}

\STATE \texttt{// Primal update (run on client $s$)}

\FOR{$s=1$ {\bfseries to} $S$}
\STATE $(\lambda_s^{k+1}, \bm{\alpha}_s^{k+1}, \hat{\bm{w}}_s^{k+1}) \leftarrow \mathfrak{C}_{s}(\bm{w}^{k+1}_s, z^{k+1}_s, \bm{\psi}^k_s, \zeta^k_s)$
\ENDFOR

\STATE \texttt{// Dual update}

\STATE $\bm{\sigma}^{k+1} \leftarrow \bm{\sigma}^k + c \cdot (\bm{t}^{k+1} - \bm{z}^{k+1} - \gamma^{k+1} \mathbf{e} - \bm{\eta}^{k+1})$

\FOR{$s=1$ {\bfseries to} $S$}
\STATE $\bm{\psi}_s^{k+1} \leftarrow \bm{\psi}_s^k + c \cdot (\bm{w}^{k+1} - \hat{\bm{w}}_s^{k+1})$

\STATE $\zeta_s^{k+1} \leftarrow \zeta_s^k + c \cdot (\rho_s\lambda_s^{k+1} + \frac{1}{N_s}\mathbf{e}^\top \bm{\alpha}_s^{k+1} - \bm{z}_s^{k+1})$
\ENDFOR

$k \leftarrow k+1$\\
\ENDWHILE
\STATE {\bfseries Output:} Solution $\bm{w}^k$
\end{algorithmic}
\end{algorithm}

\subsection*{Update for $\bm{w}$, $\bm{t}$ and $\bm{z}$}
Follow by the framework AD-LPMM, we split the primal variables into two groups and update them sequentially. For the update of the first group of variables $(\bm{w},\bm{t},\bm{z})$, we optimize the augmented Lagrangian of problem~\eqref{eq:admm_split} with a proximity term and stepsize $c > 0$ while fixing all other variables. By choosing the proximity term carefully (see Appendix~\ref{proof:add_detail_adlpmm}), this optimization problem is separable and each of subproblems can be solved analytically. Thus, the update for $(\bm{w},\bm{t},\bm{z})$ can be represented by the following operators. The primal update operator for $\bm{w}$ is defined as
\begin{equation} \label{eq:operator_w}
\mathfrak{P}_{\bm{w}}(\{\hat{\bm{w}}_s\},\{\bm{\psi}_s\}) =\frac{1}{cS} \Big (\sum_{s\in\mathcal{S}} (c \hat{\bm{w}}_s - \bm{\psi}_s ) \Big ),
\end{equation}
where the primal variables $\{\hat{\bm{w}}_s\}$ and dual variables $ \{\bm{\psi}_s\}$ are considered to be fixed when updating $\bm{w}$. Similarly, 
the primal update operator for $\bm{z}$ is defined as
\begin{equation} \label{eq:operator_z}
\mathfrak{P}_{\bm{z}}(\overline{\bm{t}},\overline{\bm{z}},\bm{\eta},\gamma,\{\lambda_s\},\{\bm{\alpha}_s\},\bm{\zeta},\bm{\sigma}) = \frac{1}{1+2S}\cdot \Big [ \bm{\pi}+\overline{\bm{t}} - \overline{\bm{z}} - \gamma \cdot \mathbf{e} - \bm{\eta} + 2S \cdot \overline{\bm{z}} + \frac{\bm{\zeta} + \bm{\sigma}}{c} \Big ].  
\end{equation}
Here, the notation $\overline{\bm{t}}$ and $\overline{\bm{z}}$ represents the values of $\bm{t}$ and $\bm{z}$ from the previous iteration, respectively. Throughout the discussion, we will consistently use the overline notation to denote the values from the previous iteration. Finally, we have
\begin{equation} \label{eq:operator_t}
\mathfrak{P}_{\bm{t}}(\overline{\bm{t}},\overline{\bm{z}},\bm{\eta},\gamma,\bm{\sigma}) = \bm{u} - \frac{\theta}{2Sc} \cdot \text{Proj}_{B_{\Vert \cdot \Vert_p}[0,1]} \left( \frac{2Sc}{\theta} \cdot \bm{u} \right),
\end{equation}
where $\bm{u} = \overline{\bm{t}} - ( \hat{\bm{q}} + \bm{\sigma} + c\cdot (\overline{\bm{t}} - \overline{\bm{z}} - \gamma \mathbf{e} - \bm{\eta})) / (2Sc)$ and $\text{Proj}_{B_{\Vert \cdot \Vert_p}[0,1]}(\cdot)$ represents the operator of orthogonal projection onto the $\ell_p$ unit ball $B_{\Vert \cdot \Vert_p}[0,1]$. It is worth noting that this projection can be computed efficiently for $p = 1,2,\infty$ (see Appendix~\ref{proof:add_detail_adlpmm}).

\subsection*{Update for $\bm{\eta}$ and $\gamma$}
For the update of the second group of variables $(\bm{\eta},\gamma,\{\lambda_s\},\{\bm{\alpha}_s\},\{\hat{\bm{w}}_s\})$, we again optimize the augmented Lagrangian of problem~\eqref{eq:admm_split} with a proximity term and stepsize $c > 0$ while fixing all other variables. Due to the specific structure of this problem, we can separately update $\bm{\eta}$, $\gamma$, and $(\{\lambda_s\},\{\bm{\alpha}_s\},\{\hat{\bm{w}}_s\})$. The primal update operator for $\bm{\eta}$ is defined as
\begin{equation} \label{eq:operator_eta}
\mathfrak{P}_{\bm{\eta}}(\bm{t},\bm{z},\overline{\bm{\eta}},\overline{\gamma},\bm{\sigma})
= \left[ \overline{\bm{\eta}} + \frac{1}{2S} \cdot \left( \frac{1}{c}\cdot \bm{\sigma} -\overline{\gamma} \cdot \mathbf{e} - \overline{\bm{\eta}} + \bm{t} - \bm{z}\right) \right]_+ .
\end{equation}

Similarly, the primal update operator for $\gamma$ is defined as
\begin{equation} \label{eq:operator_gamma}
\mathfrak{P}_{\gamma}(\bm{t},\bm{z},\overline{\bm{\eta}},\overline{\gamma},\bm{\sigma}) = \overline{\gamma} + \frac{1 + \bm{\sigma}^\top \mathbf{e} + c\cdot (- \overline{\gamma}\mathbf{e} - \overline{\bm{\eta}} + \bm{t} - \bm{z} )^\top\mathbf{e} }{2Sc} .
\end{equation}

\subsection*{Update for $\{ \lambda_s \}$, $\{ \bm{\alpha}_s \}$ and $\{ \hat{\bm{w}}_s \}$}

The update for $\{ \lambda_s \}$, $\{ \bm{\alpha}_s \}$ and $\{ \hat{\bm{w}}_s \}$ can be done separately on $(\lambda_s , \bm{\alpha}_s , \hat{\bm{w}}_s )$ for each $s\in\mathcal{S}$. For any $s\in\mathcal{S}$, we denote $\mathfrak{C}_s (\bm{w}_s,z_s,\bm{\psi}_s,\zeta_s)$ as
\begin{equation*}
\displaystyle \underset{(\lambda_s,\bm{\alpha}_s,\hat{\bm{w}}_s)\in \Omega_s} 
{\text{argmin}} \;  \zeta_s \left( \rho_s \lambda_s + \frac{1}{N_s} \mathbf{e}^\top \bm{\alpha}_{s}  \right) - \bm{\psi}_s^\top \hat{\bm{w}}_s + \frac{c}{2} \cdot \Vert \bm{w}-\hat{\bm{w}}_s \Vert_2^2 + \frac{c}{2}  \left( \rho_s \lambda_s + \frac{1}{N_s} \mathbf{e}^\top \bm{\alpha}_{s} - z_s \right)^2.
\end{equation*}
Here, the notation $\mathfrak{C}_s$ is used to indicate that this operator should be executed at the end of the $s$th client. This is because the update of $(\lambda_s , \bm{\alpha}_s , \hat{\bm{w}}_s )$ involves the set $\Omega_s$ and, consequently, the training samples $\{ (\hat{\bm{x}}_{i},\hat{y}_{i}) \}_{i=1}^{N_s}$ in client $s$.

The optimization problem associated with $\mathfrak{C}_s$ can be viewed as computing the regularized worst-case expected loss of client $s$. This can be observed from Proposition~\ref{prop:maxinf_omega}, where the problem $\inf_{(\lambda_s, \bm{\alpha}_s, \bm{w}) \in \Omega_s} \{ \rho_s \lambda_s + \mathbf{e}^\top \bm{\alpha}_s / N_s \}$ represents the worst-case expected loss of client $s$. The first term in the objective function of $\mathfrak{C}_s$ is a weighted value of that worst-case expected loss. Additionally, the second and third terms in the objective function of $\mathfrak{C}_s$ correspond to the optimization of the ``local weight parameter'' $\hat{\bm{w}}_s$. Specifically, these terms penalize significant deviations of $\hat{\bm{w}}_s$ from the ``global'' model parameter $\bm{w}$. The forth term in the objective function of $\mathfrak{C}_s$ penalizes when $\rho_s \lambda_s + \mathbf{e}^\top \bm{\alpha}_s / N_s$ deviates from $z_s$, where they should be the same for feasibility, as shown in \eqref{eq:admm_split}.

Inspired by the above observation, one can approximate $\mathfrak{C}_s (\bm{w}_s,z_s,\bm{\psi}_s,\zeta_s)$ by approximating $\Omega_s$ with a sample of data there. That is, we sample a subset of $\mathcal{I}_s$, called $\tilde{\mathcal{I}}_s$, at every iteration where $\vert \tilde{\mathcal{I}}_s \vert \ll \vert \mathcal{I}_s \vert$, then we replace $\mathcal{I}_s$ by $\tilde{\mathcal{I}}_s$ in the definition of $\mathfrak{C}_s $. The intuition behind this approximation strategy is that by using a smaller amount of data, one can estimate the worst-case expected loss of client $s$. The model parameter $\bm{w}$ is then updated based on these approximate worst-case expected losses. Since $\tilde{\mathcal{I}}_s$ is sampled at every iteration, different training samples will be used to update $\bm{w}$ at different iterations, and so the final model parameter $\bm{w}$ is still expected to be sufficiently robust. We emphasize that it is not necessary to conduct exact updates in ADMM, and the convergence rate has been proved to be consistent with $O(1/k)$ by inexact update in~\cite{bai2022inexact, xie2019si}.

\subsection*{Update for Dual Variables}
The update of dual variables is to take the gradient ascent direction in the dual problem, and this ascent direction is in fact the terms in the equality constraints \citep{beck2017first}, as shown in Algorithm~\ref{alg:adlpmm}.

\section{Numerical Experiments} \label{sec:numerical}

In this section, we compare the performances of DRFL with the standard (non-robust) approach, DRFA \citep{deng2020distributionally}, AFL \citep{mohri2019agnostic} and WAFL \citep{nguyen2022generalization} in three commonly-used datasets: heart \citep{misc_heart}, breast-cancer \citep{misc_breast_cancer_wisconsin_(original)_15}, and abalone \citep{misc_abalone_1} for SVM problem and HR problem. In this experiment, we compare the out-of-sample performance between the proposed DRFL and other approaches in SVM and HR problems, to stand for regression and classification problems, respectively. 

To assess the noise robustness of the models, we introduce different levels of Gaussian attribute noise into the training and testing data to create data heterogeneity among clients and simulate a noisy environment, respectively. We train and compare DRFL with other models in SVM problem on two scaled benchmark datasets: breast-cancer \citep{misc_breast_cancer_wisconsin_(original)_15} and heart \citep{misc_heart}, and HR problem on dataset abalone \citep{misc_abalone_1}. All optimization problems in this section are solved on an Intel 3.6 GHz processor with 32GB RAM. Except for our proposed DRFL, we also solve DRFA \citep{deng2020distributionally}, AFL \citep{mohri2019agnostic}, WAFL \citep{nguyen2022generalization} and standard approach (baseline), where standard SVM and standard  in FL setting are depicted in the followings.
\begin{corollary} \label{coro:nominal_HR}
(Standard HR) If $L$ represents the Huber loss function with threshold $\epsilon > 0$ and $\Xi = \mathbb{R}^{n+1}$, then the form of standard HR problem is
\begin{equation*}
\begin{array}{lll} 
\min & \displaystyle \frac{1}{S}\sum\limits_{s\in\mathcal{S}}\frac{1}{N_s}\sum\limits_{i\in\mathcal{I}_s} \left( \frac{1}{2}t_{si}^2 + \epsilon \cdot \alpha_{si} \right) \\
\;\! {\rm s.t.} & \displaystyle \left \langle \bm{w}, \hat{\bm{x}}_{si} \right \rangle - \hat{y}_{si} - t_{si} \leq \alpha_{si} & \forall s\in\mathcal{S}, i \in\mathcal{I}_s \\
& \displaystyle t_{si} - \left \langle \bm{w}, \hat{\bm{x}}_{si} \right \rangle + \hat{y}_{si} \leq \alpha_{si} & \forall s\in\mathcal{S}, i \in\mathcal{I}_s \\
& \displaystyle \bm{w} \in \mathbb{R}^{n}, \; \bm{t}_s, \; \bm{\alpha}_s \in \mathbb{R}^{N_s} & \forall s \in\mathcal{S}.
\end{array}
\end{equation*}
\end{corollary}

\begin{corollary} \label{coro:nominal_SVM}
(Standard SVM) If $L$ represents the hinge loss function, then the form of standard SVM problem is
\begin{align*}
\begin{array}{lll} 
\min & \displaystyle \frac{1}{S}\sum\limits_{s\in\mathcal{S}}\frac{1}{N_s} \mathbf{e}^\top \bm{\alpha}_s \\
\;\! {\rm s.t.} & \displaystyle 1-\hat{y}_{si} \left \langle \bm{w}, \hat{\bm{x}}_{si} \right \rangle \leq \alpha_{si} & \forall s\in\mathcal{S}, i \in\mathcal{I}_s \\
& \displaystyle \bm{w} \in \mathbb{R}^{n}, \; \bm{\alpha}_s \in \mathbb{R}_+^{N_s} & \forall s \in\mathcal{S}.
\end{array}
\end{align*}
\end{corollary}

Besides, we solve DRFL models with $d$ in $\ell_{1}$-norm for the sake of tractability \citep{blanchet2022optimal}. We assume two special support sets for $\bm{x}$ in the SVM problem and exhibit equivalent reformulations for $\Omega_s$ in the followings.
\begin{proposition} \label{prop:SVM1}
(SVM - Special Case~1) If $\mathcal{X} = \{ \bm{x} \in \mathbb{R}^n: - \mathbf{e} \leq \bm{x} \leq \mathbf{e} \}$, then $(\lambda_s, \bm{\alpha}_s, \bm{w}) \in \Omega_s$ is equivalent to the following constraints
\begin{equation*}
\left\{ 
\begin{array}{ll}
\displaystyle 1 + \left \langle - \hat{y}_{si}\bm{w} - \bm{\pi}_{si}^+ + \bm{\tau}_{si}^+ \;,\; \hat{\bm{x}}_{si} \right \rangle + \left \langle \bm{\pi}_{si}^+ + \bm{\tau}_{si}^+ \;,\; \mathbf{e} \right \rangle \leq \alpha_{si} & \forall i \in \mathcal{I}_s  \\
\displaystyle 1 + \left \langle \hat{y}_{si}\bm{w} - \bm{\pi}_{si}^- + \bm{\tau}_{si}^- \;,\; \hat{\bm{x}}_{si} \right \rangle + \left \langle \bm{\pi}_{si}^- + \bm{\tau}_{si}^- \;,\; \mathbf{e} \right \rangle - \kappa \cdot \lambda_s \leq \alpha_{si} & \forall i \in \mathcal{I}_s \\
\displaystyle \|\bm{\pi}_{si}^+ - \bm{\tau}_{si}^+ + \hat{y}_{si} \bm{w}\|_* \leq \lambda_s & \forall i \in \mathcal{I}_s \\
\displaystyle \|\bm{\pi}_{si}^- - \bm{\tau}_{si}^- - \hat{y}_{si} \bm{w}\|_* \leq \lambda_s & \forall i \in \mathcal{I}_s \\
\displaystyle \lambda_s \in \mathbb{R}_+, \; \bm{\alpha}_s \in \mathbb{R}_+^{N_s}, \; \bm{w} \in \mathbb{R}^n, \; \bm{\pi}_{si}^+, \; \bm{\pi}_{si}^-, \; \bm{\tau}_{si}^+, \; \bm{\tau}_{si}^- \in \mathbb{R}_+^n & \forall i \in \mathcal{I}_s.
\end{array}
\right .
\end{equation*}
\end{proposition}

\begin{proposition} \label{prop:SVM2}
(SVM - Special Case~2) If $\mathcal{X} = \{ \bm{x} \in \mathbb{R}^n: \mathbf{0} \leq \bm{x} \leq \mathbf{e} \}$, then $(\lambda_s, \bm{\alpha}_s, \bm{w}) \in \Omega_s$ is equivalent to the following constraints
\begin{equation*}
\left\{ 
\begin{array}{ll}
\displaystyle 1 + \left \langle - \hat{y}_{si}\bm{w} - \bm{\pi}_{si}^+ + \bm{\tau}_{si}^+ \;,\; \hat{\bm{x}}_{si} \right \rangle + \left \langle \bm{\pi}_{si}^+ \;,\; \mathbf{e} \right \rangle \leq \alpha_{si} & \forall i \in \mathcal{I}_s  \\
\displaystyle 1 + \left \langle \hat{y}_{si}\bm{w} - \bm{\pi}_{si}^- + \bm{\tau}_{si}^- \;,\; \hat{\bm{x}}_{si} \right \rangle + \left \langle \bm{\pi}_{si}^- \;,\; \mathbf{e} \right \rangle - \kappa \cdot \lambda_s \leq \alpha_{si} & \forall i \in \mathcal{I}_s \\
\displaystyle \|\bm{\pi}_{si}^+ - \bm{\tau}_{si}^+ + \hat{y}_{si} \bm{w}\|_* \leq \lambda_s & \forall i \in \mathcal{I}_s \\
\displaystyle \|\bm{\pi}_{si}^- - \bm{\tau}_{si}^- - \hat{y}_{si} \bm{w}\|_* \leq \lambda_s & \forall i \in \mathcal{I}_s \\
\displaystyle \lambda_s \in \mathbb{R}_+, \; \bm{\alpha}_s \in \mathbb{R}_+^{N_s}, \; \bm{w} \in \mathbb{R}^n, \; \bm{\pi}_{si}^+, \; \bm{\pi}_{si}^-, \; \bm{\tau}_{si}^+, \; \bm{\tau}_{si}^- \in \mathbb{R}_+^n & \forall i \in \mathcal{I}_s.
\end{array}
\right .
\end{equation*}
\end{proposition}

Prior to conduct this experiment, we split $60\%$ of the dataset as the training set and $40\%$ as the test set, and randomly split both sets into 3 client sample sets, \textit{i.e.}, $S=3$. Besides, we employ $5$-Fold cross-validation to select parameter values in this experiment, while accuracy and mean squared error (MSE) are used as performance metrics for classification and regression models, respectively. Meanwhile, we set a commonly used value of $\epsilon = 1.35$ for HR problem to conduct the experiment. 

\subsection{Performance Comparison Based on Class-Balanced Datasets}

To assess the robustness of models, we introduce Gaussian noise into client 1's training set, with a zero-mean and 0.5 standard deviation (SD) Gaussian noise into the heart and abalone dataset, while a 0.5 mean and 0.5 SD Gaussian noise is added into the breast-cancer dataset. Furthermore, different levels of Gaussian attribute noises are added to the test data. First, we inject Gaussian noise with fixed SD and test with different mean levels. Second, we introduce Gaussian noise with fixed mean and various SD. Third, different Gaussian noises with a certain mean and SD, which mean = ratio * SD, here ratio is a constant. 
And we tune different parameters from different sets, that is, $\rho_s \in \{1e-6, 1e-5, 1e-4, 1e-3, 1e-2, 2e-6, 2e-5, 2e-4, 2e-3, 2e-2, 5e-6, 5e-4, 5e-2, 5e-1, 1e+0 \}$, 
$\kappa \in \{1e-1, 2e-1, 3e-1, 4e-1, 5e-1, 6e-1, 7e-1, 8e-1, 9e-1, 1e+0\}$, and $\theta \in \{1e-5, 1e-4, 1e-3, 1e-2, 1e-1, 1e+0 \}$.

\begin{figure*}[!ht] %[hbt!]
\centering
\begin{minipage}{5cm}
\centering
\includegraphics[width=1\textwidth]{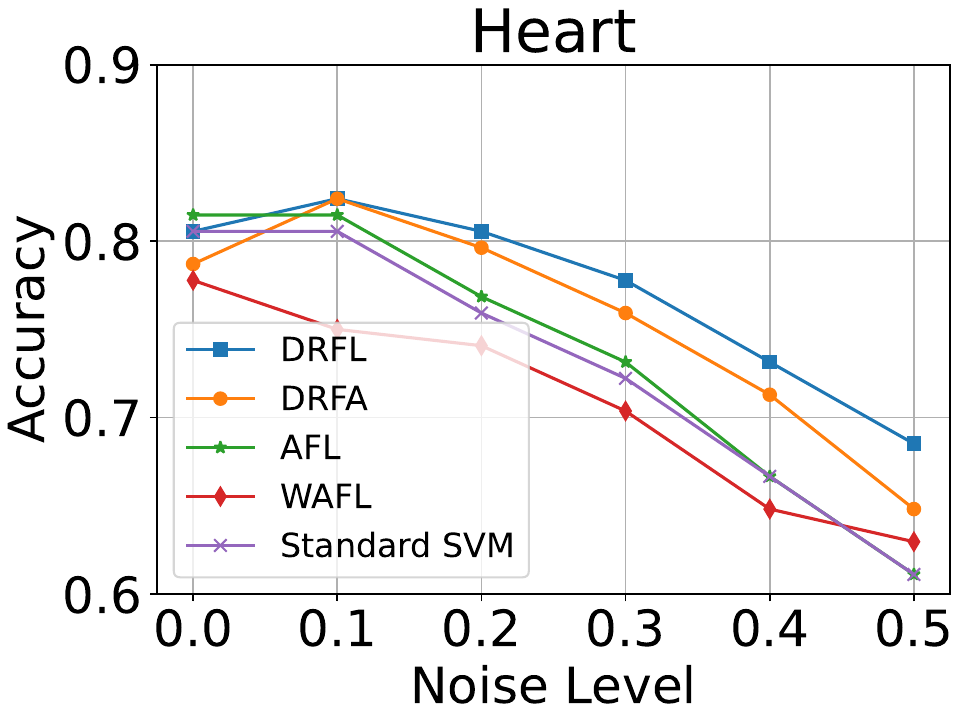}
\end{minipage}
\begin{minipage}{5cm}
\centering
\includegraphics[width=1\textwidth]{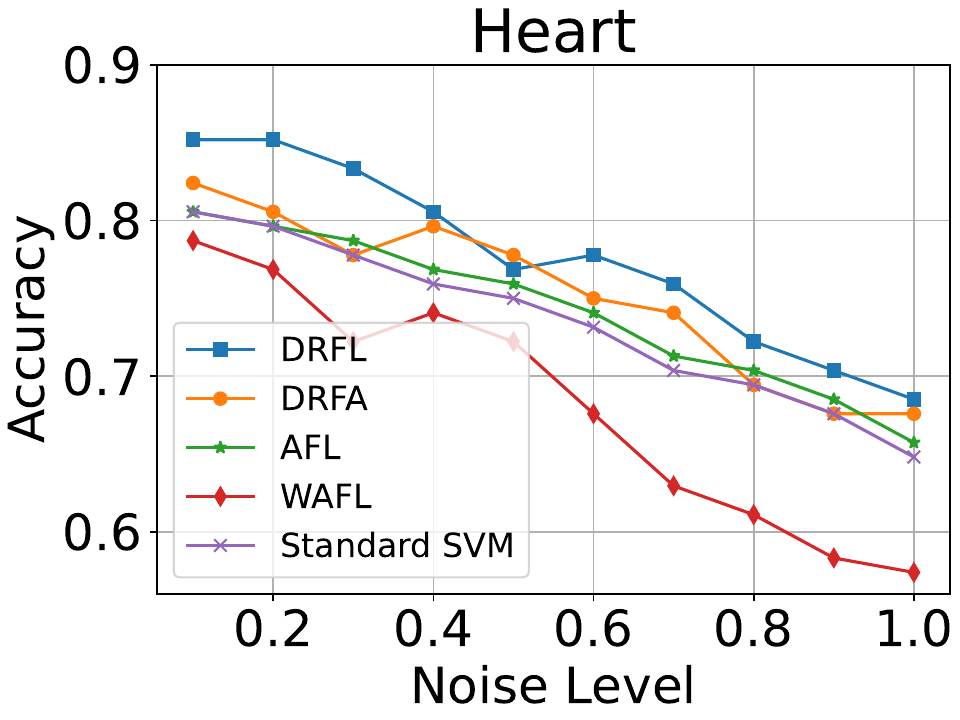}
\end{minipage}
\begin{minipage}{5cm}
\centering
\includegraphics[width=1\textwidth]{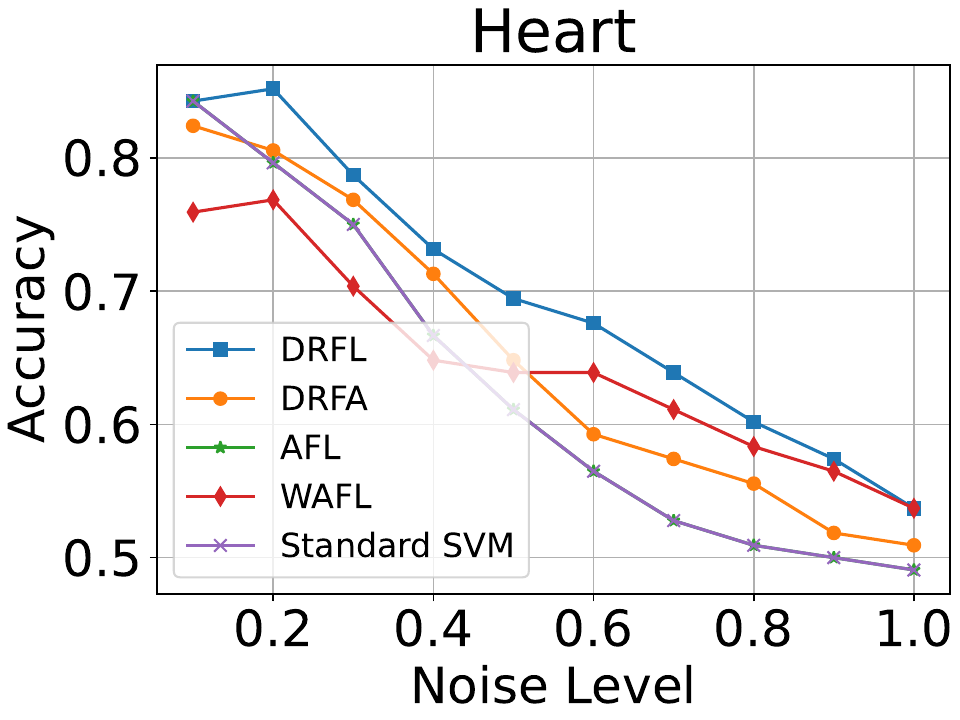}
\end{minipage}
\begin{minipage}{5cm}
\centering
\includegraphics[width=1\textwidth]{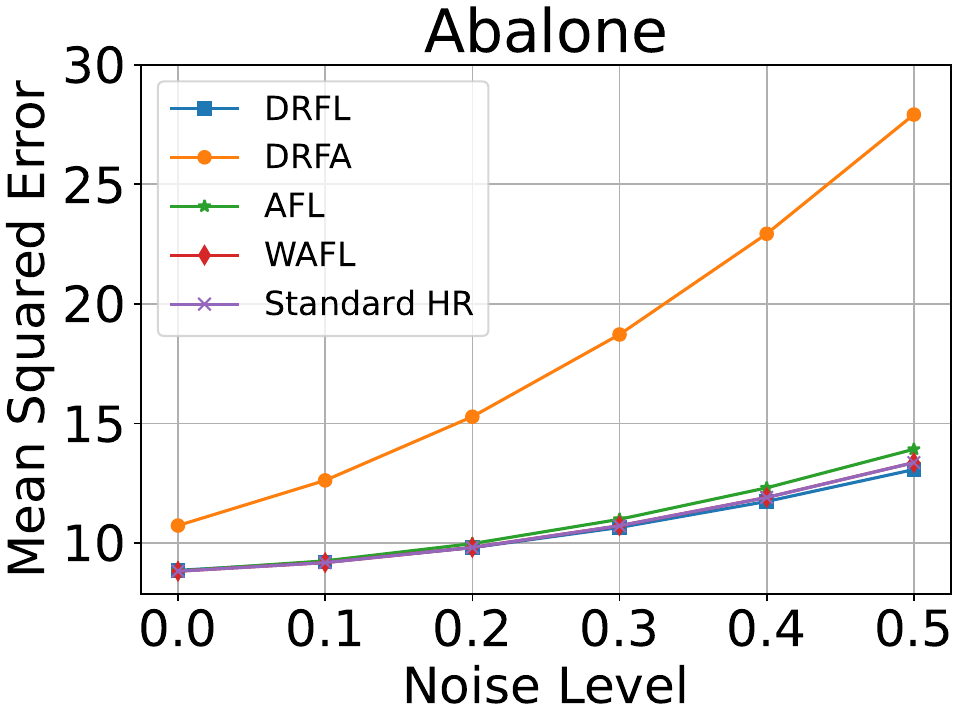}
\end{minipage}
\begin{minipage}{5cm}
\centering
\includegraphics[width=1\textwidth]{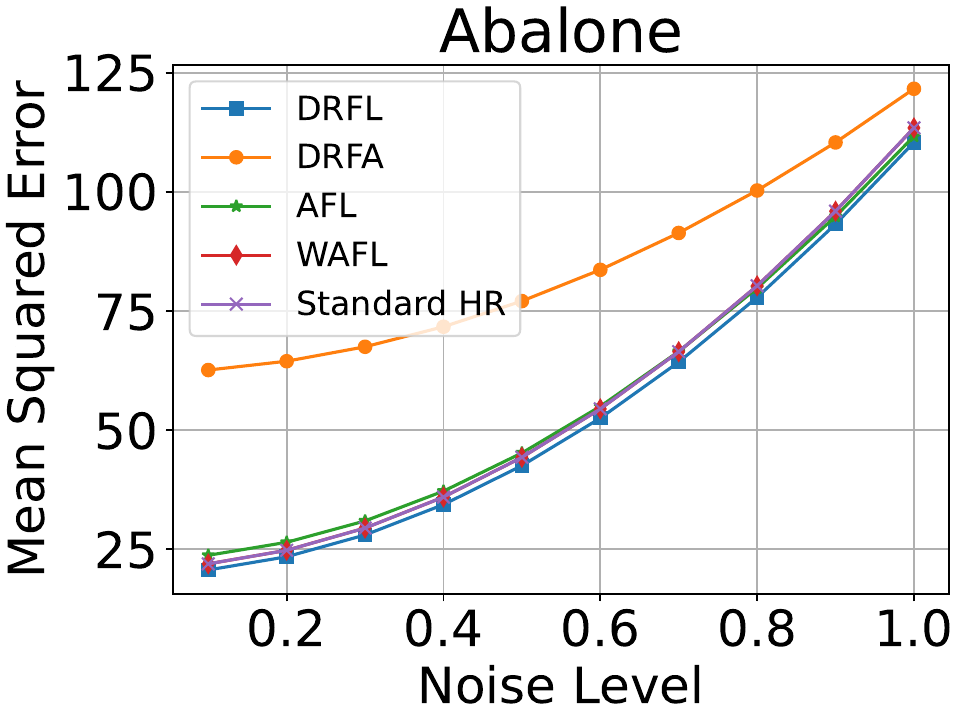}
\end{minipage}
\begin{minipage}{5cm}
\centering
\includegraphics[width=1\textwidth]{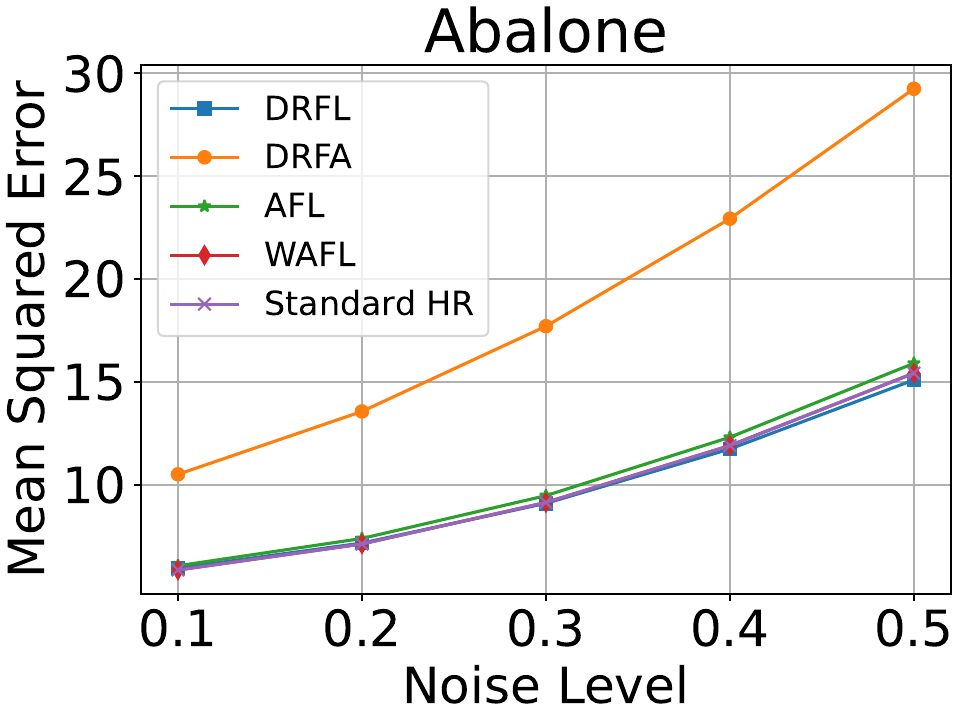}
\end{minipage}
\caption{(Heart and Abalone datasets)
Comparing noise resilience among our proposed DRFL, DRFA \citep{deng2020distributionally}, AFL \citep{mohri2019agnostic}, WAFL \citep{nguyen2022generalization} and Standard FL (baseline). Gaussian noise with increasing mean and fixed standard deviation (SD) (left). Gaussian noise with increasing SD and fixed mean (middle). Gaussian noise with mean = constant $*$ SD (right). {\bf Remark: }DRFA and standard HR exhibit equally suboptimal performance, with their results overlapping of Abalone dataset.}
\label{fig:num_perform_models}
\end{figure*}

\begin{figure*}[!ht] %[hbt!]
\centering
\begin{minipage}{5cm}
\centering
\includegraphics[width=1\textwidth]{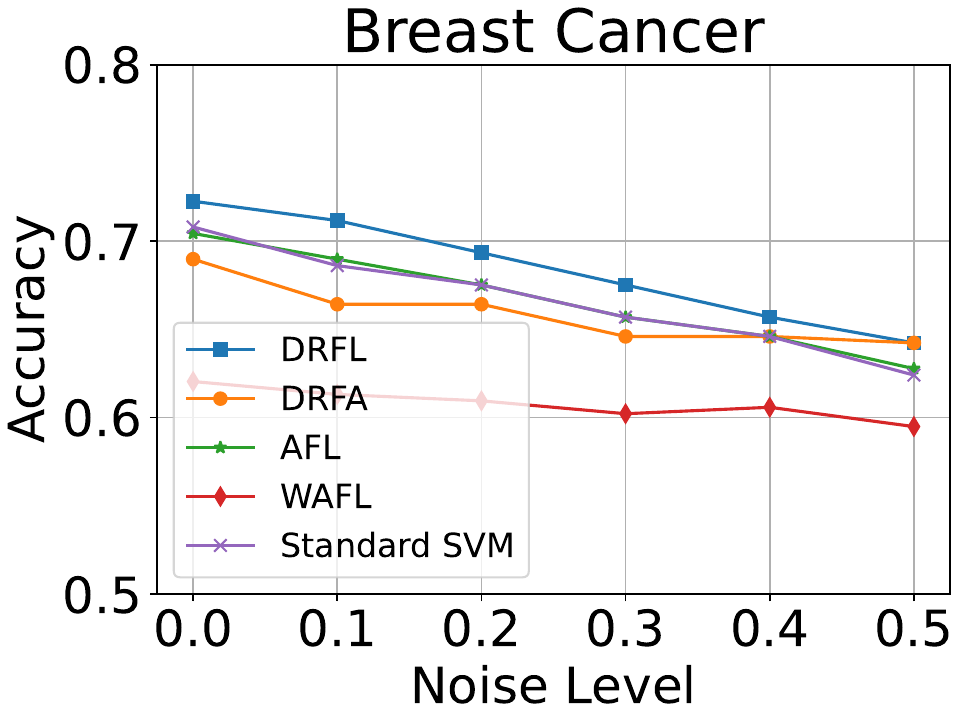}
\end{minipage}
\begin{minipage}{5cm}
\centering
\includegraphics[width=1\textwidth]{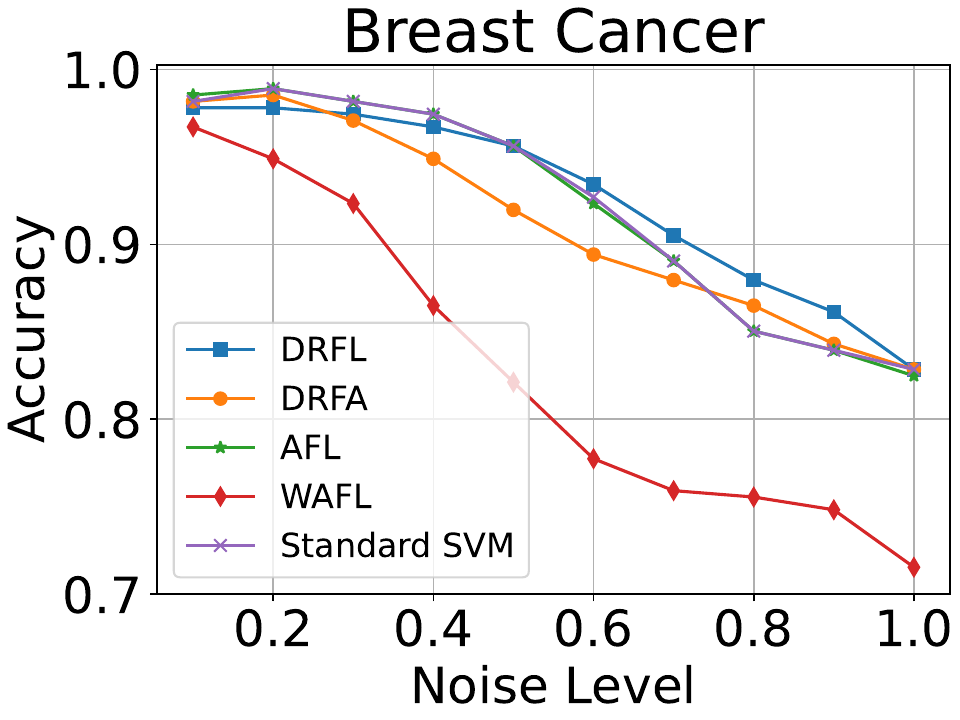}
\end{minipage}
\begin{minipage}{5cm}
\centering
\includegraphics[width=1\textwidth]{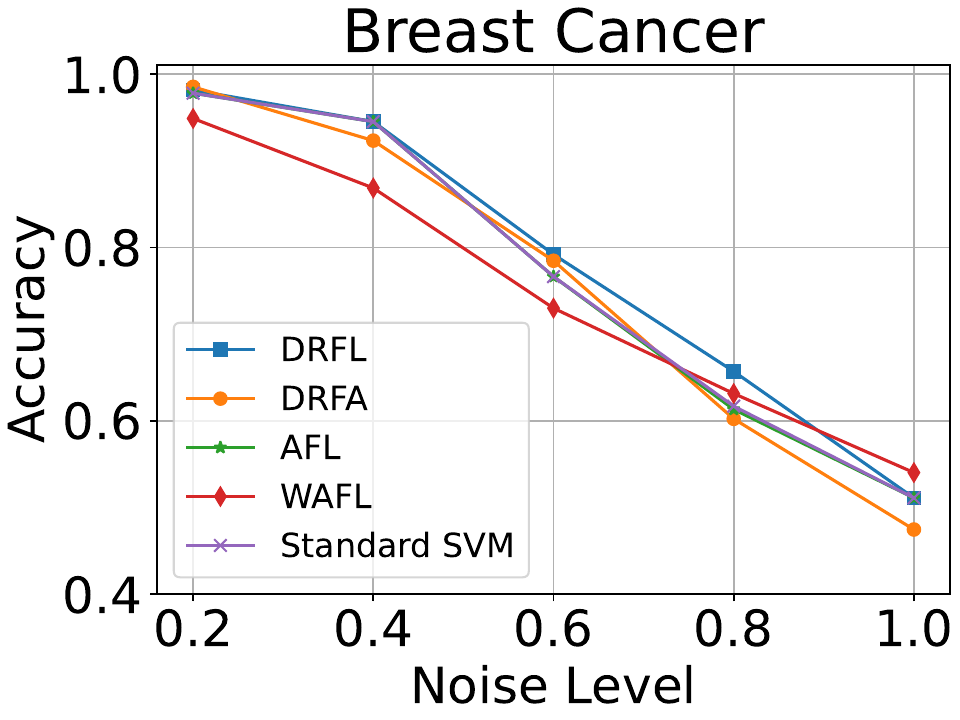}
\end{minipage}
\caption{(Breast-cancer dataset) Comparing noise resilience among our proposed DRFL, DRFA \citep{deng2020distributionally}, AFL \citep{mohri2019agnostic}, WAFL \citep{nguyen2022generalization} and Standard SVM (baseline). Gaussian noise with increasing mean and fixed SD = 2 (left). Gaussian noise with increasing SD and fixed mean = 0 (middle). Gaussian noise with mean = 2 * SD (right).}
\label{fig:num_perform_models_breast}
\end{figure*}

\newpage

Figure~\ref{fig:num_perform_models} and \ref{fig:num_perform_models_breast} show the result of noise tolerance of DRFL compared with other models. Figure~\ref{fig:num_perform_models} present results of different noise tolerance of DRFL compared with other methods in two datasets for SVM problem (figures in the first line) and HR problem (figures in the second line). Figure \ref{fig:num_perform_models_breast} demonstrates that the DRFL achieves a higher accuracy compared with other approaches in most cases, which suggests that DRFL performs well in the presence of noise, even at high levels. As we can see, DRFL outperforms other models in high-noise environments in most cases. We also test the applicability of DRFL to more complicated settings, such as class-imbalanced datasets (shown in Figure~\ref{fig:num_perform_models_breast_imbalance} and \ref{fig:num_perform_models_heart_imbalance}).

\subsection{Performance Comparison Based on Class-Imbalanced Datasets}

\begin{figure*}[!htb] %[hbt!]
\centering
\begin{minipage}{5cm}
\centering
\includegraphics[width=1\textwidth]{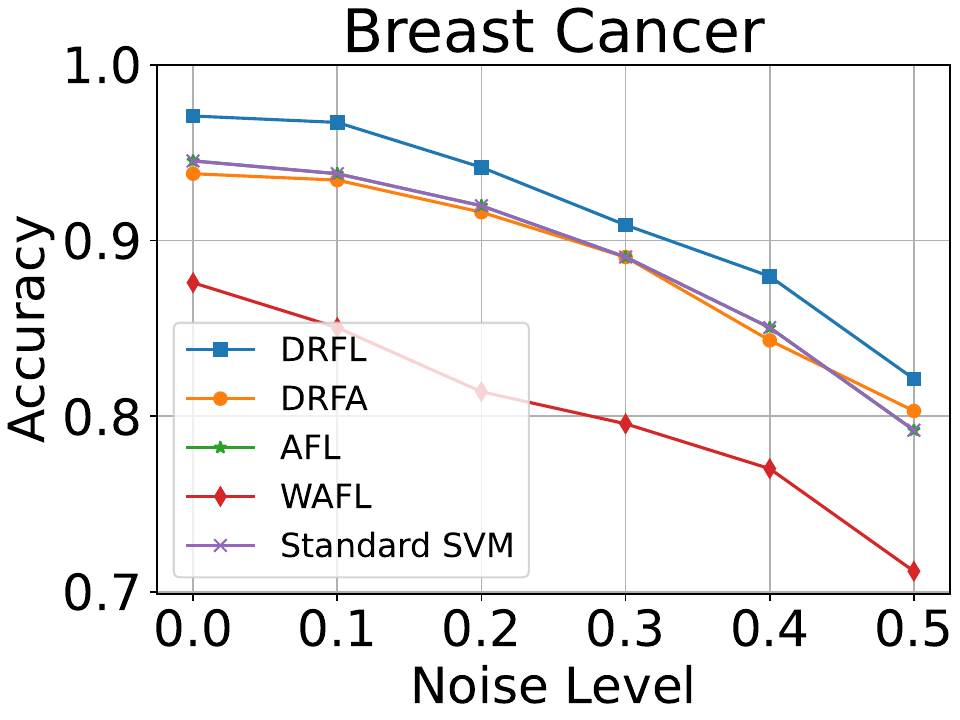}
\end{minipage}
\begin{minipage}{5cm}
\centering
\includegraphics[width=1\textwidth]{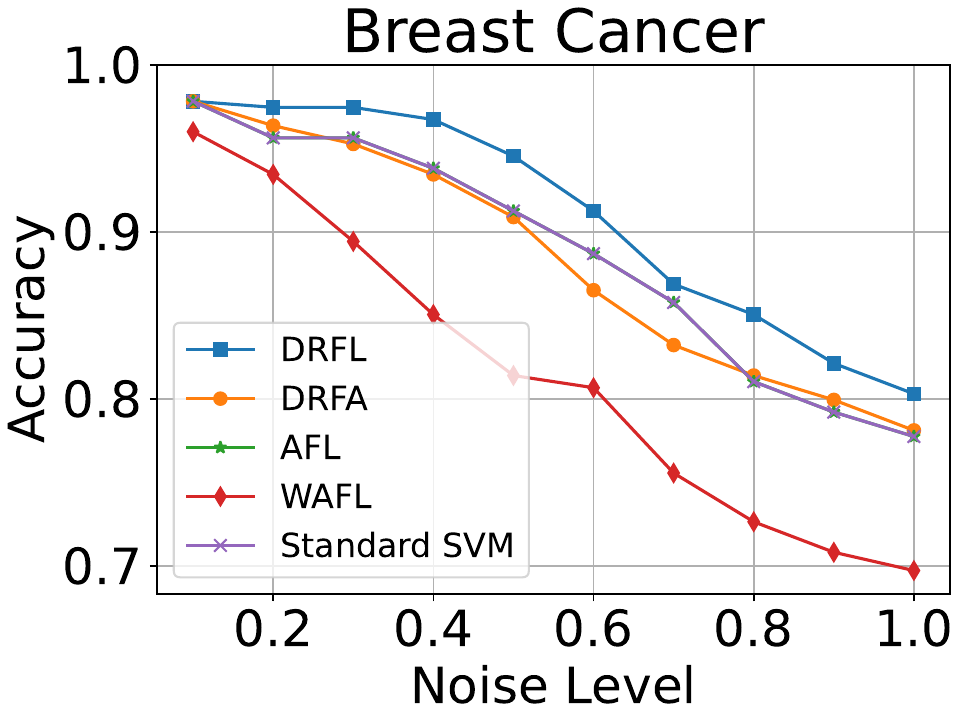}
\end{minipage}
\begin{minipage}{5cm}
\centering
\includegraphics[width=1\textwidth]{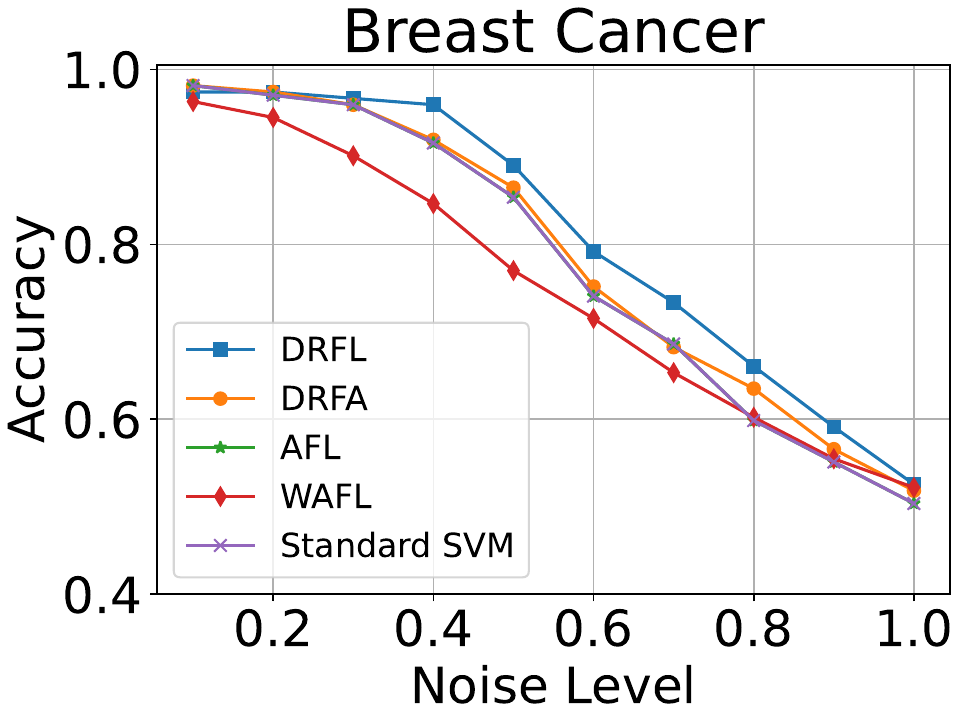}
\end{minipage}
\caption{(Breast-cancer dataset) Comparing noise resilience in imbalances class scenario among our proposed DRFL, DRFA \citep{deng2020distributionally}, AFL \citep{mohri2019agnostic}, WAFL \citep{nguyen2022generalization} and Standard SVM (baseline). Gaussian noise with increasing mean and fixed SD = 0.4 (left). Gaussian noise with increasing SD and fixed mean = 0.1 (middle). Gaussian noise with mean = 2 * SD (right).}
\label{fig:num_perform_models_breast_imbalance}
\end{figure*}

\begin{figure*}[!htb] %[hbt!]
\centering
\begin{minipage}{5cm}
\centering
\includegraphics[width=1\textwidth]{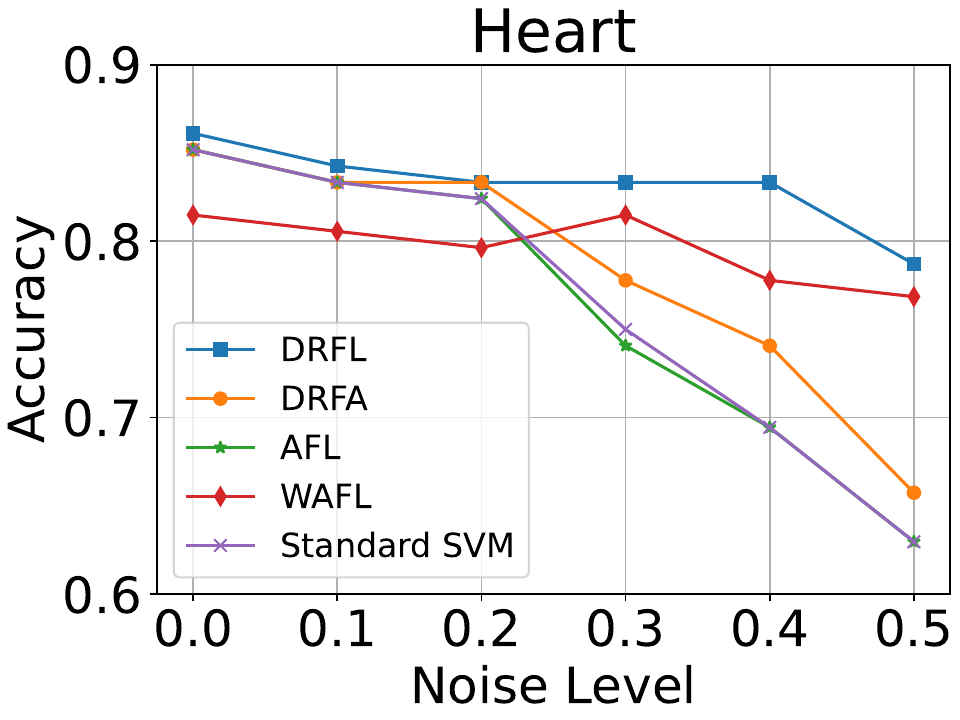}
\end{minipage}
\begin{minipage}{5cm}
\centering
\includegraphics[width=1\textwidth]{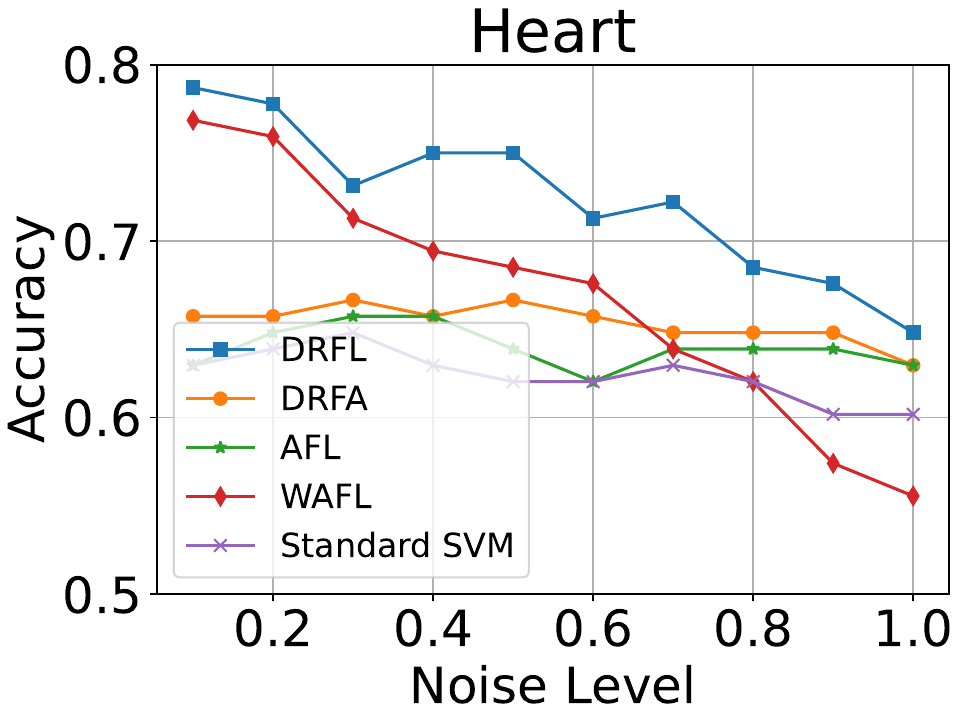}
\end{minipage}
\begin{minipage}{5cm}
\centering
\includegraphics[width=1\textwidth]
{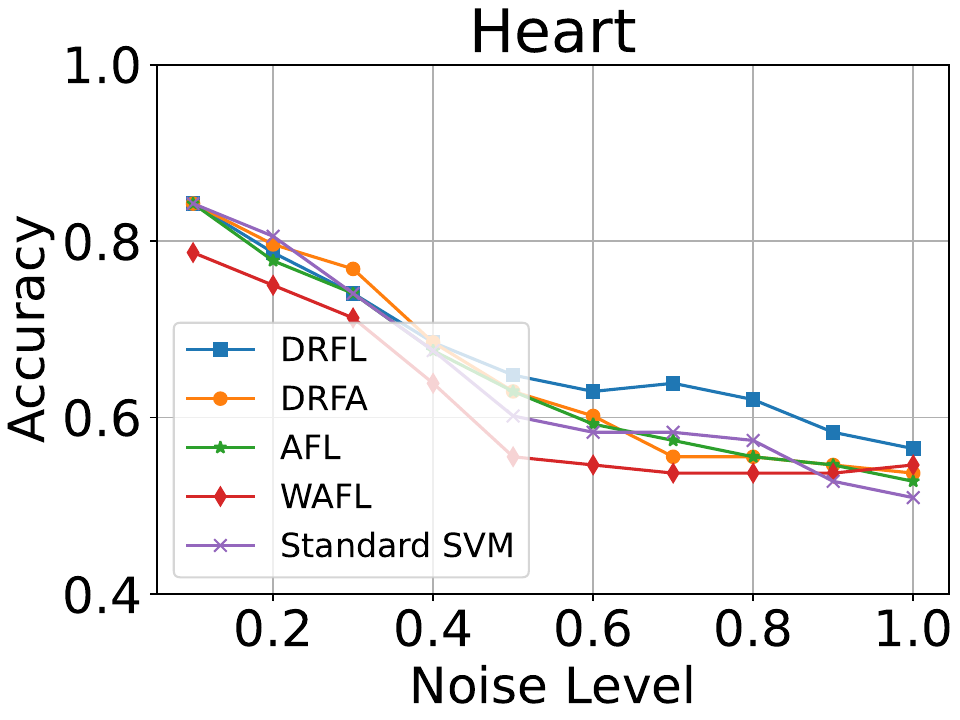}
\end{minipage}
\caption{(Heart dataset)
Comparing noise resilience in imbalance class scenario among our proposed DRFL, DRFA \citep{deng2020distributionally}, AFL \citep{mohri2019agnostic}, WAFL \citep{nguyen2022generalization} and Standard SVM (baseline). Gaussian noise with increasing mean and fixed standard deviation (SD) = 0.1 (left). Gaussian noise with increasing SD and fixed mean = 0.5 (middle). Gaussian noise with mean = 0.5*SD (right).}
\label{fig:num_perform_models_heart_imbalance}
\end{figure*}

Furthermore, we also show the results of noise tolerance of DRFL under imbalance class scenarios in Figure~\ref{fig:num_perform_models_breast_imbalance} and \ref{fig:num_perform_models_heart_imbalance}. We have partitioned the heart and breast cancer datasets into imbalanced cases. To address this, we first identify the majority class in the training set of each client. Then, we set a ratio for the sample data of the minority class. Specifically, for client 1 and 2, we randomly selected a ratio of 0.5 and for client 3 we experimented with a ratio of 1.0. The rest of the training setup remains the same as described in Section 6.1. The following results showcase the performance of the model when confronted with imbalanced classes.

\newpage

\section{Conclusion and Limitations}

We introduce DRFL to address data heterogeneity and distributional ambiguity in FL, and propose a tractable reformulation and an ADMM-based algorithm to efficiently solve our model. Numerical results depict that DRFL outperforms other models. For the limitations, we currently focus on the centralized settings with a center server, thus, the future work includes improving communication and developing algorithms based on distributed structure.

\newpage
\linespread{1}
\small

\bibliographystyle{plainnat}
\bibliography{References}

%%%%%%%%%%%%%%%%%%%%%%%%%%%%%%%%%%%%%%%%%%%%%%%%%%%%%%%%%%%%
\newpage
\linespread{1.5}
\normalsize

\appendix

\section{Technical Results and Proofs}

\noindent \textbf{Proof of Proposition~\ref{prop:maxinf_omega}.}
We first reformulate $\sup_{\mathbb{P} \in \mathcal{P}} \mathbb{E}^{\mathbb{P}}[L( f_{\bm{w}}(\tilde{\bm{x}}), \tilde{y})]$ as
\begin{equation*}
\sup_{\mathbb{P} \in \mathcal{P}} \; \mathbb{E}^{\mathbb{P}}[L( f_{\bm{w}}(\tilde{\bm{x}}), \tilde{y})] 
=
\max_{\bm{q}\in\mathcal{Q}} \sup_{\mathbb{P}_s\in\mathcal{P}_s,s\in \mathcal{S}} \sum_{s\in \mathcal{S} } q_s \cdot \mathbb{E}^{\mathbb{P}_s}[L( f_{\bm{w}}(\tilde{\bm{x}}), \tilde{y})]
=
\max_{\bm{q}\in\mathcal{Q}} \; \sum_{s\in \mathcal{S} } q_s \cdot \sup_{\mathbb{P}_s\in\mathcal{P}_s} \mathbb{E}^{\mathbb{P}_s}[L( f_{\bm{w}}(\tilde{\bm{x}}), \tilde{y})].
\end{equation*}
Denote $\bm{\xi} = (\bm{x},y)$ on $\Xi$ and by the definition of Wasserstein ball, we have
\begin{equation} \label{eq:total prob+ball defi}
\begin{array}{rl}
\displaystyle \sup_{\mathbb{P}_s\in\mathcal{P}_s} \mathbb{E}^{\mathbb{P}_s}[L( f_{\bm{w}}(\tilde{\bm{x}}), \tilde{y})]
\;\; =& \left \{
\begin{array}{ll} 
\sup\limits_{\Pi_s} \; \int_{\Xi^2} \displaystyle L( f_{\bm{w}}(\bm{x}), y) \Pi_s (d\bm{\xi},d\bm{\xi}') \\ {\rm s.t.} \ \, \int_{\Xi^2} d(\bm{\xi},\bm{\xi}') \Pi_s (d\bm{\xi},d\bm{\xi}') \leq \rho_s \\
\displaystyle \qquad \Pi_s \text{ is a joint distribution of } \bm{\xi} \text{ and } \bm{\xi}' \\
\displaystyle \qquad \text{with marginals } \mathbb{P}_s \text{ and } \hat{\mathbb{P}}_s
\end{array} 
\right.
\\
=& \left \{
\begin{array}{ll} 
\sup\limits_{\mathbb{P}_{si}} \displaystyle \frac{1}{N_s} \sum\limits_{i\in\mathcal{I}_s} \int_\Xi L( f_{\bm{w}}(\bm{x}), y) \mathbb{P}_{si}(d\bm{\xi}) \\
\displaystyle {\rm s.t.} \ \frac{1}{N_s} \sum\limits_{i\in\mathcal{I}_s} \int_\Xi d(\bm{\xi},\hat{\bm{\xi}}_{si}) \mathbb{P}_{si}(d\bm{\xi}) \leq \rho_s \\
\displaystyle \quad \ \ \int_\Xi \mathbb{P}_{si}(d\bm{\xi}) = 1, \ \forall s \in \mathcal{S},  \forall i \in \mathcal{I}_s, 
\end{array} 
\right.
\end{array}
\end{equation}
where $\mathbb{P}_{si}$ is the conditional distribution of $(\bm{x},y)$ given $(\bm{x}',y') = \hat{\bm{\xi}}_{si} = (\hat{\bm{x}}_{si},\hat{y}_{si})$ for client $s$. Using a standard duality argument, we add dual variable $\lambda_s$ and reformulate \eqref{eq:total prob+ball defi} as
\begin{equation*}
\begin{aligned}
\begin{array}{ll}
& \sup\limits_{\mathbb{P}_{si}} 
\inf\limits_{\lambda_s \geq 0} \bigg \{ \displaystyle \frac{1}{N_s} \sum\limits_{i\in\mathcal{I}_s} \int_\Xi L( f_{\bm{w}}(\bm{x}), y) \mathbb{P}_{si}(d({\bm{x},y})) + \lambda_s \Big [ \rho_s  - \frac{1}{N_s} \sum\limits_{i\in\mathcal{I}_s} \int_\Xi d((\bm{x},y),(\hat{\bm{x}}_{si},\hat{y}_{si})) \mathbb{P}_{si}(d({\bm{x},y})) \Big ] \bigg \} \\
\leq & \displaystyle \inf\limits_{\lambda_s \geq 0} \bigg \{ \rho_s\lambda_s + \sup\limits_{\mathbb{P}_{si}} \frac{1}{N_s} \sum\limits_{i\in\mathcal{I}_s}  \int_\Xi \Big [ L( f_{\bm{w}}(\bm{x}), y) - \lambda_s d((\bm{x},y),(\hat{\bm{x}}_{si},\hat{y}_{si})) \Big ] \mathbb{P}_{si}(d({\bm{x},y})) \bigg \} \\
= & \displaystyle \inf\limits_{\lambda_s \geq 0} \bigg \{ \rho_s\lambda_s + \frac{1}{N_s} \sum\limits_{i\in\mathcal{I}_s} \sup\limits_{(\bm{x},y) \in \Xi} \Big [ L( f_{\bm{w}}(\bm{x}), y) -\lambda_s  d((\bm{x},y),(\hat{\bm{x}}_{si},\hat{y}_{si})) \Big ] \bigg \},
\end{array}
\end{aligned}
\end{equation*}
where the inequality can be converted to equality since strong duality holds for any $\rho_s > 0$ due to Proposition~3.4 in \citep{shapiro2001duality}. Using epigraphical reformulation, the above problem is equivalent to
\begin{equation*}
\begin{array}{l@{\quad}ll}
\inf & \Big \{ \displaystyle \rho_s \lambda_s + \frac{1}{N_s} \mathbf{e}^\top \bm{\alpha}_s \Big \} &\\
\text{s.t.} & \displaystyle \sup_{(\bm{x},y) \in \Xi}  L( f_{\bm{w}}(\bm{x}), y) - \lambda_s  d((\bm{x},y),(\hat{\bm{x}}_{si},\hat{y}_{si})) \leq \alpha_{si} & \forall i \in \mathcal{I}_s \\
& \displaystyle \lambda_s \in \mathbb{R}_+, \; \bm{\alpha}_s \in \mathbb{R}^{N_s}, &
\end{array}
\end{equation*}
which yields to the statement in Proposition~\ref{prop:maxinf_omega}.

\begin{proposition} \label{prop:max_q dual}
The dual of the following optimization problem
\begin{equation*}
\begin{array}{l@{\quad}ll}
\max & \displaystyle \bm{z}^\top\bm{q} &\\
\text{\rm s.t.} & \displaystyle \Vert \bm{q} - \hat{\bm{q}} \Vert_p \leq \theta &  \\
& \displaystyle \mathbf{e}^\top\bm{q} = 1 &\\
& \displaystyle \bm{q} \in \mathbb{R}^S_+ &
\end{array}
\end{equation*}
is
\begin{equation*}
\begin{array}{l@{\quad}ll}
\min & \displaystyle \hat{\bm{q}}^\top(\bm{z} +\gamma  \mathbf{e} +\bm{\eta}) + \beta \cdot \theta - \gamma &\\
\text{\rm s.t.} & \displaystyle \Vert \bm{z} +\gamma  \mathbf{e} +\bm{\eta} \Vert_{p_*} \leq \beta &\\
& \displaystyle \beta \in\mathbb{R}_+, \; \gamma \in \mathbb{R}, \; \bm{\eta} \in \mathbb{R}^S_+. &
\end{array}
\end{equation*}
\end{proposition}

\noindent \textbf{Proof of Proposition~\ref{prop:max_q dual}.}
The Lagrangian dual function associated with the primal problem is
\begin{equation*}
\begin{array}{rl}
g(\beta, \gamma, \bm{\eta}) \;\;  =  &\displaystyle  \max_{\bm{q} \in \mathbb{R}^S} \bm{z}^\top\bm{q} + \beta \cdot (\theta - \Vert \bm{q} - \hat{\bm{q}} \Vert_p ) + \gamma \cdot (\mathbf{e}^\top\bm{q} - 1) + \bm{\eta}^\top\bm{q}\\
= &\displaystyle \max_{\bm{q} \in \mathbb{R}^S} \bm{q}^\top(\bm{z} +\gamma  \mathbf{e} +\bm{\eta}) - \beta \cdot \Vert \bm{q} - \hat{\bm{q}} \Vert_p + \beta \cdot  \theta  -\gamma,
\end{array}
\end{equation*}
where $\beta \in\mathbb{R}_+, \; \gamma \in \mathbb{R}$, and $\bm{\eta} \in \mathbb{R}^S_+$. By changing variables and applying convex conjugate, we have
\begin{equation*}
\begin{array}{rl}
&\displaystyle \max_{\bm{q} \in \mathbb{R}^S} \bm{q}^\top(\bm{z} +\gamma  \mathbf{e} +\bm{\eta}) - \beta \cdot \Vert \bm{q} - \hat{\bm{q}} \Vert_p + \beta \cdot  \theta  -\gamma  \\
= &\displaystyle \max_{\bm{q} \in \mathbb{R}^S} (\bm{q}- \hat{\bm{q}})^\top(\bm{z} +\gamma  \mathbf{e} +\bm{\eta}) - \beta \cdot \Vert \bm{q} - \hat{\bm{q}} \Vert_p + \hat{\bm{q}}^\top(\bm{z} +\gamma\mathbf{e} +\bm{\eta}) + \beta \cdot  \theta  -\gamma \\
= &\displaystyle
\left\{
\begin{array}{ll}
\hat{\bm{q}}^\top(\bm{z} +\gamma  \mathbf{e} +\bm{\eta}) + \beta \cdot \theta - \gamma  &  \text{if } \Vert \bm{z} +\gamma  \mathbf{e} +\bm{\eta} \Vert_{p_*} \leq \beta \\
\infty     & \text{otherwise.}
\end{array}
\right.
\end{array}
\end{equation*}
Therefore, we obtain the desired result.

\noindent \textbf{Proof of Theorem~\ref{thm:main_reform}.}
As stated in Proposition~\ref{prop:maxinf_omega}, we have
\begin{equation*}
\begin{array}{l@{\quad}l}
& \displaystyle \sup_{\mathbb{P} \in \mathcal{P}} \; \mathbb{E}^{\mathbb{P}}[L(\tilde{y} \left \langle \bm{w},\tilde{\bm{x}}\right \rangle)] \\
= & \displaystyle \max_{\bm{q}\in\mathcal{Q}} \; \sum_{s\in \mathcal{S} } q_s \cdot \inf_{(\lambda_s, \bm{\alpha}_s, \bm{w}) \in \Omega_s} \Big \{ \rho_s \lambda_s + \frac{1}{N_s} \mathbf{e}^\top \bm{\alpha}_s \Big \} \\
= & \displaystyle \inf_{(\lambda_s, \bm{\alpha}_s, \bm{w}) \in \Omega_s, s \in \mathcal{S}} \max_{\bm{q}\in\mathcal{Q}} \; \sum_{s\in \mathcal{S} } q_s \cdot \Big \{ \rho_s \lambda_s + \frac{1}{N_s} \mathbf{e}^\top \bm{\alpha}_s \Big \},
\end{array}
\end{equation*}
where the last equality holds due to minimax theorem, since the objective function is continuous and concave-convex, $\mathcal{Q}$ and $\Omega_s$ are convex, and $\mathcal{Q}$ is compact. By applying Proposition~\ref{prop:max_q dual} and the definition of $\mathcal{Q}$, we can reformulate problem~\eqref{eq:origin_problem} as
\begin{equation*}
\begin{array}{l@{\quad}ll}
\inf & \displaystyle \Big \{ \min \; \hat{\bm{q}}^\top(\bm{z} +\gamma  \mathbf{e} +\bm{\eta}) + \beta \cdot \theta - \gamma \Big \} & \\
\text{\rm s.t.} & \displaystyle \Vert \bm{z} +\gamma  \mathbf{e} +\bm{\eta} \Vert_{p_*} \leq \beta & \\
& \displaystyle z_s = \rho_s \lambda_s + \frac{1}{N_s} \mathbf{e}^\top \bm{\alpha}_s & \forall s\in\mathcal{S} \\
& \displaystyle ( \lambda_s, \bm{\alpha}_s, \bm{w})  \in \Omega_s & \forall s\in\mathcal{S} \\
& \displaystyle \bm{w}\in \mathcal{W}, \; \beta \in\mathbb{R}_+, \; \gamma \in \mathbb{R}, \; \bm{\eta} \in \mathbb{R}^S_+, \; \bm{z} \in \mathbb{R}^S, \; \lambda_s \in \mathbb{R}, \; \bm{\alpha}_s \in \mathbb{R}^{N_s} & \forall s \in \mathcal{S},
\end{array}
\end{equation*}
which is equivalent to 
\begin{equation*}
\begin{array}{l@{\quad}ll}
\inf & \displaystyle \hat{\bm{q}}^\top (\bm{z} + \bm{\eta}) + \theta \cdot \Vert \bm{z} +\gamma  \mathbf{e} +\bm{\eta} \Vert_{p_*} & \\
\text{\rm s.t.} & \displaystyle z_s = \rho_s \lambda_s + \frac{1}{N_s} \mathbf{e}^\top \bm{\alpha}_s & \forall s\in\mathcal{S}\\
& \displaystyle ( \lambda_s, \bm{\alpha}_s, \bm{w})  \in \Omega_s & \forall s\in\mathcal{S} \\
& \displaystyle \bm{w}\in \mathcal{W}, \; \gamma \in \mathbb{R}, \; \bm{\eta} \in \mathbb{R}^S_+, \; \bm{z} \in \mathbb{R}^S, \; \lambda_s \in \mathbb{R}, \; \bm{\alpha}_s \in \mathbb{R}^{N_s} & \forall s \in \mathcal{S},
\end{array}
\end{equation*}
where we merge minimum into infimum and diminish $\beta$ for simplicity. Thus we obtain the desired result.

\noindent \textbf{Proof of Proposition~\ref{prop:SVM1}.}
According to the form of~\eqref{eq:support set classification}, if $\mathcal{X} = \{ \bm{x} \in \mathbb{R}^n: - \mathbf{e} \leq \bm{x} \leq \mathbf{e} \}$, we should set $\bm{C} = [\bm{I} \; -\bm{I}]^\top$ and $\bm{d} = [\mathbf{e}^\top \; \mathbf{e}^\top]^\top$,
because $-\mathbf{e} \leq \bm{x} \leq \mathbf{e}$ is equivalent to the following two inequalities
\begin{equation*}
\displaystyle \bm{x} \leq \mathbf{e} \ \ \text{and} \ - \bm{x} \leq \mathbf{e}.
\end{equation*}
We split $\bm{\phi}_{si}^+$ into two variables, $\bm{\pi}_{si}^+$ and $ \bm{\tau}_{si}^+$, while split $\bm{\phi}_{si}^-$ into $\bm{\pi}_{si}^-$ and $ \bm{\tau}_{si}^-$ in the same way. Then we have
\begin{equation*}
\bm{C}^\top \bm{\phi}_{si}^+ = \bm{\pi}_{si}^+ - \bm{\tau}_{si}^+,
\end{equation*}
and
\begin{equation*}
\bm{C}^\top \bm{\phi}_{si}^- = \bm{\pi}_{si}^- - \bm{\tau}_{si}^-.
\end{equation*}
Also, we can derive
\begin{equation*}
\left \langle \bm{\phi}_{si}^+,\bm{d}-\bm{C} \hat{\bm{x}}_{si} \right \rangle = \left \langle \bm{\pi}_{si}^+ \;,\; \mathbf{e}-\bm{x} \right \rangle + \left \langle \bm{\tau}_{si}^+ \;,\; \mathbf{e}+\bm{x} \right \rangle,
\end{equation*}
and
\begin{equation*}
\left \langle \bm{\phi}_{si}^-,\bm{d}-\bm{C} \hat{\bm{x}}_{si} \right \rangle = \left \langle \bm{\pi}_{si}^- \;,\; \mathbf{e}-\bm{x} \right \rangle + \left \langle \bm{\tau}_{si}^- \;,\; \mathbf{e}+\bm{x} \right \rangle.
\end{equation*}
By substituting the above equations into the reformulation in Corollary~\ref{coro:SVM}, we can obtain the desired result.

\noindent \textbf{Proof of Proposition~\ref{prop:SVM2}.}
According to the form of~\eqref{eq:support set classification}, if $\mathcal{X} = \{ \bm{x} \in \mathbb{R}^n: \mathbf{0} \leq \bm{x} \leq \mathbf{e} \}$, then we should set $\bm{C} = [\bm{I} \; -\bm{I}]^\top$ and $\bm{d} = [\mathbf{e}^\top \; \mathbf{0}^\top]^\top$,
because $\mathbf{0} \leq \bm{x} \leq \mathbf{e}$ is equivalent to the following two inequalities
\begin{equation*}
\displaystyle \bm{x} \leq \mathbf{e} \ \ \text{and} \ - \bm{x} \leq \mathbf{0}.
\end{equation*}
We split $\bm{\phi}_{si}^+$ into two variables, $\bm{\pi}_{si}^+$ and $ \bm{\tau}_{si}^+$, while split $\bm{\phi}_{si}^-$ into $\bm{\pi}_{si}^-$ and $ \bm{\tau}_{si}^-$ in the same way. Then we have
\begin{equation*}
\bm{C}^\top \bm{\phi}_{si}^+ = \bm{\pi}_{si}^+ - \bm{\tau}_{si}^+,
\end{equation*}
and
\begin{equation*}
\bm{C}^\top \bm{\phi}_{si}^- = \bm{\pi}_{si}^- - \bm{\tau}_{si}^-.
\end{equation*}
Also, we can derive
\begin{equation*}
\left \langle \bm{\phi}_{si}^+,\bm{d}-\bm{C} \hat{\bm{x}}_{si} \right \rangle = \left \langle \bm{\pi}_{si}^+ \;,\; \mathbf{e}-\bm{x} \right \rangle + \left \langle \bm{\tau}_{si}^+ \;,\; \bm{x} \right \rangle,
\end{equation*}
and
\begin{equation*}
\left \langle \bm{\phi}_{si}^-,\bm{d}-\bm{C} \hat{\bm{x}}_{si} \right \rangle = \left \langle \bm{\pi}_{si}^- \;,\; \mathbf{e}-\bm{x} \right \rangle + \left \langle \bm{\tau}_{si}^- \;,\; \bm{x} \right \rangle.
\end{equation*}
By substituting the above equations into the reformulation in Corollary~\ref{coro:SVM}, we can obtain the desired result.

\noindent \textbf{Proof of Corollary~\ref{coro:nominal_HR}.}
As shown in Section~\ref{sec:prelim}, the standard federated learning problem is in the form of problem~\eqref{eq:non_robust_problem}. If $L$ represents the Huber loss function, then problem~\eqref{eq:non_robust_problem} becomes
\begin{equation*}
\begin{array}{ll}
& \displaystyle \min_{\bm{w} \in \mathcal{W}} \; \frac{1}{S} \sum_{s\in\mathcal{S}} \frac{1}{N_s} \sum_{i \in \mathcal{I}_s} L_{\rm R}(\left \langle \bm{w} , \hat{\bm{x}}_{si} \right \rangle - \hat{y}_{si}) \\
= & \displaystyle \min_{\bm{w} \in \mathcal{W}} \; \frac{1}{S} \sum_{s\in\mathcal{S}} \frac{1}{N_s} \sum_{i \in \mathcal{I}_s} \min_{t_{si}} \Big \{ \frac{1}{2}t_{si}^2 + \epsilon \cdot \vert \left \langle \bm{w}, \hat{\bm{x}}_{si} \right \rangle - \hat{y}_{si} - t_{si} \vert \Big \},
\end{array}
\end{equation*}
where the equality holds due to the equivalent reformulation of Huber loss function \citep{shafieezadeh2019regularization}. We then transform the above problem into the epigraphic form by adding variables $\{\alpha_{si}\}$, which leads to the following form
\begin{equation*}
\begin{array}{lll} 
\min & \displaystyle \frac{1}{S}\sum\limits_{s\in\mathcal{S}}\frac{1}{N_s}\sum\limits_{i\in\mathcal{I}_s} \left( \frac{1}{2}t_{si}^2 + \epsilon \cdot \alpha_{si} \right) \\
\;\! {\rm s.t.} & \displaystyle \vert \left \langle \bm{w}, \hat{\bm{x}}_{si} \right \rangle - \hat{y}_{si} - t_{si} \vert \leq \alpha_{si} & \forall s\in\mathcal{S}, i \in\mathcal{I}_s \\
& \displaystyle \bm{w} \in \mathcal{W}, \; \bm{t}_s, \; \bm{\alpha}_s \in \mathbb{R}^{N_s} & \forall s \in\mathcal{S}.
\end{array}
\end{equation*}
To delete the absolute operator, we split the inequality constraints into two equivalent constraints
\begin{equation*}
\left \{
\begin{array}{ll}
\displaystyle \left \langle \bm{w}, \hat{\bm{x}}_{si} \right \rangle - \hat{y}_{si} - t_{si} \leq \alpha_{si} & \forall s \in\mathcal{S}, i \in \mathcal{I}_s \\
\displaystyle t_{si} - \left \langle \bm{w}, \hat{\bm{x}}_{si} \right \rangle + \hat{y}_{si} \leq \alpha_{si} & \forall s \in\mathcal{S}, i \in \mathcal{I}_s,
\end{array}
\right.
\end{equation*}
and obtain the desired result.

\noindent \textbf{Proof of Corollary~\ref{coro:nominal_SVM}.}
As shown in Section~\ref{sec:prelim}, the standard federated learning problem is in the form of problem~\eqref{eq:non_robust_problem}. If $L$ represents the hinge loss function, then problem~\eqref{eq:non_robust_problem} becomes
\begin{equation*}
\begin{array}{ll}
& \displaystyle \min_{\bm{w} \in \mathcal{W}} \; \frac{1}{S} \sum_{s\in\mathcal{S}} \frac{1}{N_s} \sum_{i \in \mathcal{I}_s} L_{\rm C}(\hat{y}_{si} \left \langle \bm{w} , \hat{\bm{w}}_{si} \right \rangle) \\
= & \displaystyle \min_{\bm{w} \in \mathcal{W}} \; \frac{1}{S} \sum_{s\in\mathcal{S}} \frac{1}{N_s} \sum_{i \in \mathcal{I}_s} \max \big \{ 0, 1 - \hat{y}_{si} \left \langle \bm{w}, \hat{\bm{x}}_{si} \right \rangle \big \},
\end{array}
\end{equation*}
We then transform the above problem into the epigraphic form by adding variables $\{\alpha_{si}\}$, which leads to the following form
\begin{equation*}
\begin{array}{lll} 
\min & \displaystyle \frac{1}{S} \sum\limits_{s\in\mathcal{S}}\frac{1}{N_s} \mathbf{e}^\top \bm{\alpha}_s \\
\;\! {\rm s.t.} & \displaystyle \max \big \{ 0, 1 - \hat{y}_{si} \left \langle \bm{w}, \hat{\bm{x}}_{si} \right \rangle \big \} \leq \alpha_{si} & \forall s\in\mathcal{S}, i \in\mathcal{I}_s \\
& \displaystyle \bm{w} \in \mathcal{W}, \; \bm{\alpha_s} \in \mathbb{R}^{N_s} & \forall s \in\mathcal{S}.
\end{array}
\end{equation*}
To delete the maximum operator, we split the inequality constraints into two equivalent constraints
\begin{equation*}
\left \{
\begin{array}{ll}
\displaystyle 0 \leq \alpha_{si} & \forall s \in\mathcal{S}, i \in \mathcal{I}_s \\
\displaystyle 1 - \hat{y}_{si} \left \langle \bm{w}, \hat{\bm{x}}_{si} \right \rangle \leq \alpha_{si} & \forall s \in\mathcal{S}, i \in \mathcal{I}_s,
\end{array}
\right.
\end{equation*}
and obtain the desired result.

\section{Reformulations for DRFL for Different Machine Learning Models} \label{appendix:loss_funcs}

\subsection{Different Definitions for Metric \texorpdfstring{$d$}{}} \label{appendix:metric_d}

We assume transportation metric $d$ has different definitions in different applications. In this paper, we mainly discuss regression and classification problems, so we denote $d_{\rm R}$ and $d_{\rm C}$ as the metric for these two problems, respectively, and show the definitions as
\begin{equation} \label{d_reg}
    d_{\rm R}((\bm{x},y),(\hat{\bm{x}},\hat{y})) = \Vert \bm{x}-\hat{\bm{x}} \Vert + \kappa \cdot \vert y - \hat{y} \vert,
\end{equation}
and
\begin{equation} \label{d_class}
    d_{\rm C}((\bm{x},y),(\hat{\bm{x}},\hat{y})) = \|\bm{x}-\hat{\bm{x}} \| + \kappa \cdot \mathbb{I}_{\{{y} \neq \hat{y}\}},
\end{equation}
for some $\kappa > 0$. Here, $\mathbb{I}_{\{\cdot\}}$ represents an indicator function. Equation of $d_{\rm R}(\cdot)$ and $d_{\rm C}(\cdot)$ represent the distance metric for regression and classification problems, respectively. In equation~\eqref{d_reg}, $\Vert \bm{x}-\hat{\bm{x}}\Vert$ represents the distance (in norm) between the input vectors $\bm{x}$ and $\hat{\bm{x}}$, which measures the dissimilarity between the input features, and $\vert y - \hat{y} \vert$ means the absolute difference between the output values $y$ and $\hat{y}$, which quantifies the dissimilarity between the target outputs. Here, $\kappa$ denotes the cost associated with shifting probability mass across the output space in regression problem. A larger value of $\kappa$ places more emphasis on minimizing the output difference. 

In equation~\eqref{d_class}, $\mathbb{I}_{\{\cdot\}}$ represents an indicator function, which equals to 1 if the output labels $y$ and $\hat{y}$ are different, and 0 otherwise. It serves as a penalty term when the labels are not the same. Here, $\kappa$ quantifies the penalty incurred when changing a label $y$ in classification problems. A larger value of $\kappa$ increases the penalty for misclassification.

As we mentioned before, $\Omega_s$ can be simplified into different forms based on different structures of $L$, $f_{\bm{w}}$ and $d$. In the following, we refer to some definitions of $L$, $f_{\bm{w}}$ and $d$, as well as results in prior work \citep{shafieezadeh2019regularization} to demonstrate the simplifications for $\Omega_s$.

\subsection{Simplifications for \texorpdfstring{$\Omega_s$}{} (Regression Models)}
In regression models, the output $y$ is continuous and $\mathcal{Y} = \mathbb{R}$. We assume that loss functions are in the form of $L(f_{\bm{w}}(\bm{x}), y) = L_{\rm R}(\left \langle \bm{w}, \bm{x} \right \rangle - y)$ and $\Xi$ is convex and closed. In the following, we explicitly demonstrate $\Omega_s$ for Huber, $\epsilon$-intensive and pinball loss functions. The model parameter $\bm{w}$ is in $\mathcal{W} = \mathbb{R}^n$. We also assume that the support set $\Xi$ can be represented as
\begin{equation} \label{eq:support set regression}
\Xi = \{ (\bm{x}, y) \in \mathbb{R}^{n+1}: \bm{C}_1 \bm{x} + \bm{c}_2 y \leq \bm{d}, \bm{C}_1 \in \mathbb{R}^{r \times n}, \bm{c}_2 \in \mathbb{R}^r, \bm{d} \in \mathbb{R}^r \},
\end{equation}
for $\epsilon$-intensive and pinball loss functions under the assumption that there exists a Slater point $(\bm{x}_S, y_S) \in \mathbb{R}^{n+1}$ with $\bm{C}_1 \bm{x}_S + \bm{c}_2 y_S < \bm{d}$.

\begin{corollary} 
(HR) If $L$ represents the Huber loss function with threshold $\epsilon \geq 0$ and $\Xi = \mathbb{R}^{n+1}$, then $( \lambda_s, \bm{\alpha}_s, \bm{w}) \in \Omega_s $ is equivalent to the following constraints
\begin{equation*}
\left\{ 
\begin{array}{ll}
\displaystyle 1/2 \mu_{si}^2 + \epsilon \vert \left \langle \bm{w}, \hat{\bm{x}}_{si} \right \rangle - \hat{y}_{si} - \mu_{si} \vert \leq \alpha_{si} & \forall i \in \mathcal{I}_s  \\
\displaystyle \epsilon \| (\bm{w}, -1) \|_* \leq \lambda_s & \\
\displaystyle \lambda_s \in \mathbb{R}_+, \; \bm{\alpha}_s \in \mathbb{R}^{N_s}, \; \bm{w} \in \mathbb{R}^n, \; \bm{\mu}_s \in \mathbb{R}^{N_s}. &
\end{array}
\right.
\end{equation*}
In particular, due to the convexity of above equivalent constraints, we can conclude that both problem~\eqref{eq:main_theorem} and~\eqref{eq:admm_split} are convex and solvable in this setting.
\end{corollary}

\begin{corollary}
(SVR) If $L$ represents the $\epsilon$-intensive loss function for some $\epsilon \geq 0$ and $\Xi$ is in the form of~\eqref{eq:support set regression}, then $( \lambda_s, \bm{\alpha}_s, \bm{w})  \in \Omega_s $ is equivalent to the following constraints
\begin{equation*}
\left\{ 
\begin{array}{ll}
\displaystyle \hat{y}_{si} - \left \langle \bm{w}, \hat{\bm{x}}_{si} \right \rangle - \epsilon + \left \langle \bm{\phi}_{si}^+, \bm{d} - \bm{C}_1 \hat{\bm{x}}_{si} - \bm{c}_2 \hat{y}_{si} \right \rangle \leq \alpha_{si} & \forall i \in \mathcal{I}_s  \\
\displaystyle \left \langle \bm{w}, \hat{\bm{x}}_{si} \right \rangle - \hat{y}_{si} - \epsilon + \left \langle \bm{\phi}_{si}^-, \bm{d} - \bm{C}_1 \hat{\bm{x}}_{si} - \bm{c}_2 \hat{y}_{si} \right \rangle \leq \alpha_{si} & \forall i \in \mathcal{I}_s \\
\displaystyle \| (\bm{C}_1^\top \bm{\phi}_{si}^+ + \bm{w}, \bm{c}_2^\top \bm{\phi}_{si}^+ - 1) \|_* \leq \lambda_s & \forall i \in \mathcal{I}_s \\
\displaystyle \| (\bm{C}_1^\top \bm{\phi}_{si}^- - \bm{w}, \bm{c}_2^\top \bm{\phi}_{si}^- + 1) \|_* \leq \lambda_s & \forall i \in \mathcal{I}_s \\
\displaystyle \lambda_s \in \mathbb{R}_+, \; \bm{\alpha}_s \in \mathbb{R}_+^{N_s}, \; \bm{w} \in \mathbb{R}^n, \; \bm{\phi}_{si}^+, \; \bm{\phi}_{si}^- \in \mathbb{R}_+^r & \forall i \in \mathcal{I}_s.
\end{array}
\right .
\end{equation*}
In particular, due to the convexity of above equivalent constraints, we can conclude that both problem~\eqref{eq:main_theorem} and~\eqref{eq:admm_split} are convex and solvable in this setting.
\end{corollary}

\begin{corollary}
(Quantile Regression) If $L$ represents the pinball loss function for some $\epsilon \in [0, 1]$ and $\Xi$ is in the form of~\eqref{eq:support set regression}, then $( \lambda_s, \bm{\alpha}_s, \bm{w}) \in \Omega_s $ is equivalent to the following constraints
\begin{equation*}
\left\{ 
\begin{array}{ll}
\displaystyle \epsilon (\hat{y}_{si} - \left \langle \bm{w}, \hat{\bm{x}}_{si} \right \rangle) + \left \langle \bm{\phi}_{si}^+, \bm{d} - \bm{C}_1 \hat{\bm{x}}_{si} - \bm{c}_2 \hat{y}_{si} \right \rangle \leq \alpha_{si} & \forall i \in \mathcal{I}_s  \\
\displaystyle (1 - \epsilon) (\left \langle \bm{w}, \hat{\bm{x}}_{si} \right \rangle - \hat{y}_{si}) + \left \langle \bm{\phi}_{si}^-, \bm{d} - \bm{C}_1 \hat{\bm{x}}_{si} - \bm{c}_2 \hat{y}_{si} \right \rangle \leq \alpha_{si} & \forall i \in \mathcal{I}_s \\
\displaystyle \| (\bm{C}_1^\top \bm{\phi}_{si}^+ + \epsilon \bm{w}, \bm{c}_2^\top \bm{\phi}_{si}^+ - \epsilon) \|_* \leq \lambda_s & \forall i \in \mathcal{I}_s \\
\displaystyle \| (\bm{C}_1^\top \bm{\phi}_{si}^- - (1 - \epsilon) \bm{w}, \bm{c}_2^\top \bm{\phi}_{si}^- + 1 - \epsilon) \|_* \leq \lambda_s & \forall i \in \mathcal{I}_s \\
\displaystyle \lambda_s \in \mathbb{R}_+, \; \bm{\alpha}_s \in \mathbb{R}_+^{N_s}, \; \bm{w} \in \mathbb{R}^n, \; \bm{\phi}^+_{si}, \; \bm{\phi}^-_{si} \in \mathbb{R}_+^r & \forall i \in \mathcal{I}_s.
\end{array}
\right .
\end{equation*}
In particular, due to the convexity of above equivalent constraints, we can conclude that both problem~\eqref{eq:main_theorem} and~\eqref{eq:admm_split} are convex and solvable in this setting.
\end{corollary}

\subsection{Simplifications for \texorpdfstring{$\Omega_s$}{} (Classification Models)}
In classification models, the output $y$ is categorical and $\mathcal{Y} = \{ -1, \; +1\}$. We assume that loss functions are in the form of $L(f_{\bm{w}}(\bm{x}), y) = L_{\rm C}(y \left \langle \bm{w}, \bm{x} \right \rangle)$ and $\mathcal{X}$ is convex and closed. In the following, we explicitly demonstrate $\Omega_s$ for hinge loss, smooth hinge loss and logloss functions. The model parameter $\bm{w}$ is in $\mathcal{W} = \mathbb{R}^n$. We also assume that the support set $\mathcal{X}$ can be represented as
\begin{equation} \label{eq:support set classification}
\mathcal{X} = \{ \bm{x} \in \mathbb{R}^n: \bm{C} \bm{x} \leq \bm{d}, \bm{C} \in \mathbb{R}^{r \times n}, \bm{d} \in \mathbb{R}^r \},
\end{equation}
for the hinge loss function under the assumption that $\mathcal{X}$ admits a Slater point $\bm{x}_S \in \mathbb{R}^n$ with $\bm{C} \bm{x}_S < \bm{d}$.

\begin{corollary} \label{coro:SVM}
(SVM) If $L$ represents the hinge loss function, and the support set $\mathcal{X}$ is in the form of~\eqref{eq:support set classification}, then $( \lambda_s, \bm{\alpha}_s, \bm{w}) \in \Omega_s $ is equivalent to the following constraints
\begin{equation*}
\left\{ 
\begin{array}{ll}
\displaystyle 1 - \hat{y}_{si} \left \langle \bm{w},\hat{\bm{x}}_{si} \right \rangle + \left \langle \bm{\phi}_{si}^+,\bm{d}-\bm{C} \hat{\bm{x}}_{si} \right \rangle \leq \alpha_{si} & \forall i \in \mathcal{I}_s  \\
\displaystyle 1 + \hat{y}_{si} \left \langle \bm{w},\hat{\bm{x}}_{si} \right \rangle + \left \langle \bm{\phi}_{si}^-,\bm{d}-\bm{C} \hat{\bm{x}}_{si} \right \rangle - \kappa \cdot \lambda_s \leq \alpha_{si} & \forall i \in \mathcal{I}_s \\
\displaystyle \|\bm{C}^\top \bm{\phi}_{si}^+ + \hat{y}_{si} \bm{w}\|_* \leq \lambda_s & \forall i \in \mathcal{I}_s \\
\displaystyle \|\bm{C}^\top \bm{\phi}_{si}^- - \hat{y}_{si} \bm{w}\|_* \leq \lambda_s & \forall i \in \mathcal{I}_s \\
\displaystyle \lambda_s \in \mathbb{R}_+, \; \bm{\alpha}_s \in \mathbb{R}_+^{N_s}, \; \bm{w} \in \mathbb{R}^n, \; \bm{\phi}_{si}^+, \; \bm{\phi}_{si}^- \in \mathbb{R}_+^r & \forall i \in \mathcal{I}_s.
\end{array}
\right .
\end{equation*}
In particular, due to the convexity of above equivalent constraints, we can conclude that both problem~\eqref{eq:main_theorem} and~\eqref{eq:admm_split} are convex and solvable in this setting.
\end{corollary}

\begin{corollary}
(SVM with smooth hinge loss) If $L$ represents the smooth hinge loss function and $\mathcal{X} = \mathbb{R}^n$, then $( \lambda_s, \bm{\alpha}_s, \bm{w}) \in \Omega_s $ is equivalent to the following constraints
\begin{equation*}
\left\{ 
\begin{array}{ll}
\displaystyle 1/2 \cdot (\phi^+_{si} - \hat{y}_{si} \left \langle \bm{w}, \hat{\bm{x}}_{si} \right \rangle)^2 + \pi^+_{si} \leq \alpha_{si} & \forall i \in \mathcal{I}_s  \\
\displaystyle 1/2 \cdot (\phi^-_{si} + \hat{y}_{si} \left \langle \bm{w}, \hat{\bm{x}}_{si} \right \rangle)^2 + \pi^-_{si} - \kappa \cdot \lambda_s \leq \alpha_{si} & \forall i \in \mathcal{I}_s \\
\displaystyle 1 - \phi^+_{si} \leq \pi^+_{si}, \; 1 - \phi^-_{si} \leq \pi^-_{si} & \forall i \in \mathcal{I}_s \\
\displaystyle \| \bm{w} \|_* \leq \lambda_s & \\
\displaystyle \lambda_s \in \mathbb{R}_+, \; \bm{\alpha}_s \in \mathbb{R}^{N_s}, \; \bm{w} \in \mathbb{R}^n, \; \bm{\phi}^+_s, \; \bm{\phi}^-_s \in \mathbb{R}^{N_s}, \; \bm{\pi}^+_s, \; \bm{\pi}^-_s \in \mathbb{R}_+^{N_s}. &
\end{array}
\right.
\end{equation*}
In particular, due to the convexity of above equivalent constraints, we can conclude that both problem~\eqref{eq:main_theorem} and~\eqref{eq:admm_split} are convex and solvable in this setting.
\end{corollary}

\begin{corollary}
(Logistics Regression) If $L$ represents the logloss function and $\mathcal{X} = \mathbb{R}^n$, then $( \lambda_s, \bm{\alpha}_s, \bm{w})  \in \Omega_s $ is equivalent to the following constraints
\begin{equation*}
\left\{ 
\begin{array}{ll}
\displaystyle \log \big (1+\exp(-\hat{y}_{si}\left \langle \bm{w},\hat{\bm{x}}_{si}\right \rangle) \big ) \leq \alpha_{si} & \forall i \in \mathcal{I}_s  \\
\displaystyle \log \big (1+\exp(\hat{y}_{si}\left \langle \bm{w},\hat{\bm{x}}_{si}\right \rangle) \big ) -\kappa \cdot \lambda_s \leq \alpha_{si} & \forall i \in \mathcal{I}_s \\
\displaystyle \| \bm{w} \|_* \leq \lambda_s & \\
\displaystyle \lambda_s \in \mathbb{R}_+, \; \bm{\alpha}_s \in \mathbb{R}^{N_s}, \; \bm{w} \in \mathbb{R}^n. &
\end{array}
\right .
\end{equation*}
In particular, due to the convexity of above equivalent constraints, we can conclude that both problem~\eqref{eq:main_theorem} and~\eqref{eq:admm_split} are convex and solvable in this setting.
\end{corollary}

\section{Reformulations for Existed Models}
\subsection{AFL}
When $\rho_s = 0$ for all $s \in \mathcal{S}$, DRFL reduces to AFL~\citep{mohri2019agnostic}, then we have
\begin{equation} \label{eq:reform_rho_0}
\begin{array}{ll}
\displaystyle \inf_{\bm{w} \in \mathcal{W}}\; \sup_{\mathbb{P} \in \mathcal{P}}\; \mathbb{E}^\mathbb{P} [ L( f_{\bm{w}}(\tilde{\bm{x}}), \tilde{y}) ] & \displaystyle = \inf_{\bm{w} \in \mathcal{W}} \; \max_{\bm{q} \in \mathcal{Q}} \; \sum_{s\in\mathcal{S}} q_s \cdot \mathbb{E}^{\hat{\mathbb{P}}_s} [ L( f_{\bm{w}}(\bm{x}), y) ] \\
& \displaystyle = \inf_{\bm{w} \in \mathcal{W}} \; \max_{\bm{q} \in \mathcal{Q}} \; \sum_{s\in\mathcal{S}} q_s \cdot \frac{1}{N_s} \sum\limits_{i\in\mathcal{I}_s}   L( f_{\bm{w}}(\hat{\bm{x}}_{si}),\hat{y}_{si}).
\end{array}
\end{equation}
By using the trick in Proposition~\ref{prop:max_q dual}, problem~\eqref{eq:reform_rho_0} is equivalent to
\begin{equation} \label{eq:main_theorem_AFL}
\begin{array}{l@{\quad}ll}
\inf & \displaystyle \hat{\bm{q}}^\top (\bm{z} + \bm{\eta}) + \theta \cdot \Vert \bm{z} +\gamma  \mathbf{e} +\bm{\eta} \Vert_{p_*} & \\
\text{\rm s.t.} & \displaystyle z_s = \frac{1}{N_s} \mathbf{e}^\top \bm{\alpha}_s & \forall s\in\mathcal{S} \\
& \displaystyle  L( f_{\bm{w}}(\hat{\bm{x}}_{si}), \hat{y}_{si}) \leq \alpha_{si}  & \forall s\in\mathcal{S}, i \in\mathcal{I}_s \\
& \displaystyle \bm{w}\in \mathcal{W}, \; \gamma \in \mathbb{R}, \; \bm{\eta} \in \mathbb{R}^S_+, \; \bm{z} \in \mathbb{R}^S, \; \bm{\alpha}_s \in \mathbb{R}^{N_s} & \forall s \in \mathcal{S},
\end{array}
\end{equation}

\subsubsection{For SVM}
The loss function for SVM is
\begin{equation*}
L( f_{\bm{w}}(\bm{x}), y) = \max \{ 0, 1 - (y \left \langle \bm{w} , \bm{x} \right \rangle) \},
\end{equation*}
thus we derive the optimization problem~\eqref{eq:main_theorem_AFL} as
\begin{equation} \label{eq:reform_main_AFL_DRFA}
\begin{array}{rll}
\inf & \displaystyle \hat{\bm{q}}^\top (\bm{z} + \bm{\eta}) + \theta \cdot \Vert \bm{z} +\gamma  \mathbf{e} +\bm{\eta} \Vert_{p_*} & \\
\text{\rm s.t.} & \displaystyle z_s = \frac{1}{N_s} \mathbf{e}^\top \bm{\alpha}_s & \forall s\in\mathcal{S} \\ & \displaystyle 1-\hat{y}_{si} \left \langle \bm{w}, \hat{\bm{x}}_{si} \right \rangle \leq \alpha_{si} & \forall s\in\mathcal{S}, i \in\mathcal{I}_s \\
& \displaystyle \bm{w}\in \mathcal{W}, \; \gamma \in \mathbb{R}, \; \bm{\eta} \in \mathbb{R}^S_+, \; \bm{z} \in \mathbb{R}^S, \; \bm{\alpha}_s \in \mathbb{R}^{N_s}_+ & \forall s \in \mathcal{S}.
\end{array}
\end{equation}

\subsubsection{For Huber Regression}
The loss function for Huber Regression is
\begin{equation*} 
L( f_{\bm{w}}(\bm{x}), y) = \left\{
\begin{array}{ll}
\displaystyle (\left \langle \bm{w} , \bm{x} \right \rangle - y)^2 / 2 & \text{if } \vert \left \langle \bm{w} , \bm{x} \right \rangle - y \vert \leq \epsilon \\
\displaystyle \epsilon (\vert \left \langle \bm{w} , \bm{x} \right \rangle - y \vert - \epsilon / 2) & \text{otherwise,}
\end{array}
\right.
\end{equation*}
thus problem~\eqref{eq:main_theorem_AFL} becomes
\begin{equation} 
\begin{array}{rll}
\inf & \displaystyle \hat{\bm{q}}^\top (\bm{z} + \bm{\eta}) + \theta \cdot \Vert \bm{z} +\gamma  \mathbf{e} +\bm{\eta} \Vert_{p_*} & \\
\text{\rm s.t.} & \displaystyle z_s = \frac{1}{N_s} \mathbf{e}^\top \bm{\alpha}_s & \forall s\in\mathcal{S} \\ 
&\displaystyle 1/2 \mu_{si}^2 + \epsilon \vert \left \langle \bm{w}, \hat{\bm{x}}_{si} \right \rangle - \hat{y}_{si} - \mu_{si} \vert \leq \alpha_{si} & \forall s\in\mathcal{S}, i \in\mathcal{I}_s \\
& \displaystyle \bm{w} \in \mathbb{R}^{n}, \; \gamma \in \mathbb{R}, \; \bm{\eta} \in \mathbb{R}^S_+, \; \bm{z} \in \mathbb{R}^S, \; \bm{\mu}_s, \bm{\alpha}_s \in \mathbb{R}^{N_s} & \forall s \in\mathcal{S}.
\end{array}
\end{equation}

\subsection{DRFA}
Suppose $\rho_s = 0$ and $\mathcal{Q} = \Delta_S$, DRFL recudes to DRFA~\citep{deng2020distributionally}, and Problem~\eqref{eq:origin_problem} can also be rewriten as problem~\eqref{eq:reform_rho_0}.
Recall that $\Delta_S = \{\bm{q}\in\mathbb{R}^S_+ ~|~ \mathbf{e}^\top \bm{q} = 1 \}$, then the dual of the following optimization problem
\begin{equation*}
\begin{array}{l@{\quad}ll}
\max & \displaystyle \bm{z}^\top\bm{q} &\\
\text{\rm s.t.} & \displaystyle \mathbf{e}^\top\bm{q} = 1 &\\
& \displaystyle \bm{q} \in \mathbb{R}^S_+ &
\end{array}
\end{equation*}
is
\begin{equation*} \label{eq:dual_Delta_S}
\begin{array}{l@{\quad}ll}
\min & \displaystyle  \gamma \\ 
\text{\rm s.t.} & \displaystyle  \bm{z} -\gamma  \mathbf{e} \leq 0 \\
& \displaystyle \gamma \in \mathbb{R} 
\end{array}
\end{equation*}
Finally we reformulate the optimization problem~\eqref{eq:reform_rho_0} as
\begin{equation} \label {eq:main_theorem_DRFA}
\begin{array}{rll}
\inf & \displaystyle \gamma & \\
\text{\rm s.t.} & \displaystyle z_s = \frac{1}{N_s} \mathbf{e}^\top \bm{\alpha}_s & \forall s\in\mathcal{S} \\ 
& \displaystyle  z_s -\gamma \leq 0 & \forall s\in\mathcal{S}\\
& \displaystyle  L( f_{\bm{w}}(\hat{\bm{x}}_{si}), \hat{y}_{si}) \leq \alpha_{si}  & \forall s\in\mathcal{S}, i \in\mathcal{I}_s \\
& \displaystyle \bm{w} \in \mathbb{R}^{n}, \; \gamma \in \mathbb{R}, \; \bm{\alpha}_s \in \mathbb{R}^{N_s} & \forall s \in\mathcal{S}.
\end{array}
\end{equation}

\subsubsection{For SVM}
By replacing $L( f_{\bm{w}}(\hat{\bm{x}}_{si}), \hat{y}_{si})$ by the corresponding loss function of SVM, problem~$\eqref{eq:main_theorem_DRFA}$ becomes
\begin{equation} 
\begin{array}{rll}
\inf & \displaystyle \gamma & \\
\text{\rm s.t.} & \displaystyle z_s = \frac{1}{N_s} \mathbf{e}^\top \bm{\alpha}_s & \forall s\in\mathcal{S} \\ 
& \displaystyle  z_s -\gamma \leq 0 & \forall s\in\mathcal{S}\\
& \displaystyle 1-\hat{y}_{si} \left \langle \bm{w}, \hat{\bm{x}}_{si} \right \rangle \leq \alpha_{si} & \forall s\in\mathcal{S}, i \in\mathcal{I}_s \\
& \displaystyle \bm{w} \in \mathbb{R}^{n}, \; \gamma \in \mathbb{R}, \; \bm{\alpha}_s \in \mathbb{R}_+^{N_s} & \forall s \in\mathcal{S}.
\end{array}
\end{equation}

\subsubsection{For Huber Regression}
By replacing $L( f_{\bm{w}}(\hat{\bm{x}}_{si}), \hat{y}_{si})$ by the corresponding loss function of Huber Regression, problem~$\eqref{eq:main_theorem_DRFA}$ becomes
\begin{equation} 
\begin{array}{rll}
\inf & \displaystyle \gamma & \\
\text{\rm s.t.} & \displaystyle z_s = \frac{1}{N_s} \mathbf{e}^\top \bm{\alpha}_s & \forall s\in\mathcal{S} \\ 
& \displaystyle  z_s -\gamma \leq 0 & \forall s\in\mathcal{S}\\
&\displaystyle 1/2 \mu_{si}^2 + \epsilon \vert \left \langle \bm{w}, \hat{\bm{x}}_{si} \right \rangle - \hat{y}_{si} - \mu_{si} \vert \leq \alpha_{si} & \forall s\in\mathcal{S}, i \in\mathcal{I}_s \\
& \displaystyle \bm{w} \in \mathbb{R}^{n}, \; \gamma \in \mathbb{R}, \; \bm{\mu}_s, \bm{\alpha}_s \in \mathbb{R}^{N_s} & \forall s \in\mathcal{S}.
\end{array}
\end{equation}

\section{Additional Details on Example~\ref{example:volume_compare}} \label{appendix:example_problem_setting}
Consider a simple system containing two clients with distributions $\mathbb{P}_1^\star$ and $\mathbb{P}_2^\star$ and their weights $q_1$ and $q_2$, respectively. In this example, we assume $\mathbb{P}_s^*$ is a discrete distribution, that is $\mathbb{P}_s(x = x_{si}) = p_{si}$, $i \in \{1, 2, 3\}$, $\bm{p}_s \in \Delta_3$. For simplicity, we assume each client has the same Wasserstein radius $\rho$, and tune $\rho$ from $0.00001$ to $0.5$ with step $0.00001$.

In the first stage, we examine the probability of containing the underlying distribution $\mathbb{P}^\star = q_1 \cdot \mathbb{P}_1^\star + q_2 \cdot \mathbb{P}_2^\star$, that is, $P(\mathbb{P}^\star \subseteq \mathbb{P})$. Here are the four main steps:
\begin{enumerate}
\item Generate 1000 samples from $\mathbb{P}^\star$. For each sample generation, we first randomly sample the client type from $\{1, 2\}$ by the probability $q_1$ and $q_2$, respectively. Then we sample data value for client $s$ from $\{ x_{s1}, x_{s2}, x_{s3}\}$ according to probability $\{ p_{s1}, p_{s2}, p_{s3}\}$, respectively.

\item Establish an empirical distribution $\hat{\mathbb{P}}_{agg} = \hat{q}_1 \cdot \hat{\mathbb{P}}_1 + \hat{q}_2 \cdot \hat{\mathbb{P}}_2$ from the generated samples by calculating $\hat{q}_s$ and $\{ \hat{p}_{s1}, \hat{p}_{s2}, \hat{p}_{s3}\}$, $s = \{1, 2\}$.

\item For a given $\rho$, check if $\hat{\mathbb{P}}_{agg}$ contains $\mathbb{P}^\star$, that is to check $W(\mathbb{P}^\star, \hat{\mathbb{P}}_{agg}) \leq \rho$ for WAFL and $W(\mathbb{P}_s^*, \hat{\mathbb{P}}_s) \leq \rho$ for DRFL. Tag satisfied $\hat{\mathbb{P}}_{agg}$.

\item Repeat step 1-3 for 100 times for each value of $\rho$ and calculate the proportion of satisfied $\hat{\mathbb{P}}_{agg}$, \textit{i.e.}, $P(\mathbb{P}^\star \subseteq \mathbb{P})$.
\end{enumerate}

In the second stage, we compare the distribution volume for the same guarantee level. Before that, we conclude a list of $\rho$, where each component in the list corresponds to the smallest $\rho$ to achieve different levels of $P(\mathbb{P}^\star \subseteq \mathbb{P})$ in the first stage. We consider the probability levels from 0.1 to 1 with step 0.05. Here are four main steps:
\begin{enumerate}
\item Generate 1000 samples from $\mathbb{P}^\star$ and obtain a fixed empirical distribution $\hat{\mathbb{P}}_f$ by taking step 1-2 in the first stage.

\item Generate a random distribution $\mathbb{P}'$ by randomly sampling $q_s$ and $\{ p_{s1}, p_{s2}, p_{s3}\}$ from $(0,1)$, and make sure $q_1 + q_2 = 1$ as well as $\sum_{i = 1}^3 p_{si} = 1$ for $s = \{1, 2\}$.

\item For a given probability level and the corresponding smallest $\rho$, check if $\mathbb{P}'$ is inside $\hat{\mathbb{P}}_f$, that is to check $W(\mathbb{P}', \hat{\mathbb{P}}_f) \leq \rho$ for WAFL, and $W(\mathbb{P}_s', \hat{\mathbb{P}}_{fs}) \leq \rho$ for DRFL. Tag satisfied $\mathbb{P}'$.

\item Repeat step 2-3 for 10,000 times for different probability level, and calculate the normalized distribution volume, which equals to (amount of satisfied $\mathbb{P}'$ / 10,000). 
\end{enumerate}

\section{Additional Details on AD-LPMM Algorithm} \label{proof:add_detail_adlpmm}
Since $(\lambda_s, \bm{\alpha}_s, \bm{w}) \in \Omega_s$ can not be represented as linear constraints in $\lambda_s$, $\bm{\alpha}_s$, and $\bm{w}$, we apply the spirit of AD-LPMM algorithm to split the decision variables of reformulation~\eqref{eq:admm_split} into two groups and update them separately. The proposed algorithm is summarized in Algorithm~\ref{alg:adlpmm}. Firstly, we introduce the augmented Lagrangian 
\begin{equation*}
\begin{array}{rl}
& \displaystyle \mathcal{L}(\bm{w},\bm{t},\bm{z},\bm{\eta},\gamma,\{\lambda_s\},\{\bm{\alpha}_s\},\{\hat{\bm{w}}_s\};\{\bm{\psi}_s\},\bm{\zeta},\bm{\sigma}) \\
= & \displaystyle \hat{\bm{q}}^\top\bm{t} + \theta \cdot \Vert \bm{t} \Vert_{p_*} - \gamma + \bm{\sigma}^\top (\bm{t} - \bm{z} - \gamma \mathbf{e} - \bm{\eta}) + \sum_{s\in\mathcal{S}} \zeta_s \left( \rho_s \lambda_s + \frac{1}{N_s} \mathbf{e}^\top \bm{\alpha}_{s} - z_s\right) \\
& \displaystyle + \sum_{s\in\mathcal{S}} \bm{\psi}_s^\top (\bm{w}-\hat{\bm{w}}_s) + \frac{c}{2} \Vert \bm{t} - \bm{z} - \gamma \mathbf{e} - \bm{\eta} \Vert^2 + \frac{c}{2} \sum_{s\in\mathcal{S}} \left( \rho_s \lambda_s + \frac{1}{N_s} \mathbf{e}^\top \bm{\alpha}_{s} - z_s \right)^2 \\
& \displaystyle + \frac{c}{2} \sum_{s\in\mathcal{S}} \Vert \bm{w}-\hat{\bm{w}}_s \Vert^2.
\end{array}
\end{equation*}
To implement the splitting strategy, we will update two groups of $(\bm{w},\bm{t},\bm{z})$ and $(\bm{\eta},\gamma,\{\lambda_s\},\{\bm{\alpha}_s\},\{\hat{\bm{w}}_s\})$ one by one.

\subsection{Adding Proximal Term for Updating \texorpdfstring{$(\bm{w},\bm{t},\bm{z})$}{}}
The optimization problem associated with the update for the first group $(\bm{w},\bm{t},\bm{z})$ has the objective function
\begin{equation*}
\displaystyle \hat{\bm{q}}^\top\bm{t} + \theta \cdot \Vert \bm{t} \Vert_{p_*} + \bm{\sigma}^\top (\bm{t} - \bm{z}) + \sum_{s\in\mathcal{S}} \bm{\psi}_s^\top \bm{w} - \bm{\zeta}^\top \bm{z} + \frac{c}{2} \Vert \bm{t} - \bm{z} - \gamma \mathbf{e} - \bm{\eta} \Vert^2 + \frac{c}{2} \sum_{s\in\mathcal{S}} \Vert \bm{w}-\hat{\bm{w}}_s \Vert^2 + \frac{c}{2} \Vert \bm{z} -\bm{\pi}\Vert_2^2,
\end{equation*}
where $\bm{\pi} \in \mathbb{R}^S$ with $\pi_s = \rho_s \lambda_s + \frac{1}{N_s} \mathbf{e}^\top \bm{\alpha}_{s} $. To efficiently compute the update, we add a proximity term as in Alternating Direction Proximal Method of Multipliers (AD-PMM). In particular, we denote $\bm{v}_1^\top = [\bm{t}^\top \; \bm{z}^\top ]$, and we have
\begin{equation*}
\begin{array}{ll}
\displaystyle \Vert \bm{t} - \bm{z} - \gamma \mathbf{e} - \bm{\eta} \Vert^2 & = \Vert \bm{t} - \bm{z} \Vert_2^2 - 2(\bm{t} - \bm{z} )^\top(\gamma \mathbf{e} + \bm{\eta}) + \Vert \gamma \mathbf{e} + \bm{\eta} \Vert_2^2 \\
& \displaystyle = \Vert \bm{A}_1\bm{v}_1 \Vert_2^2 - 2(\bm{A}_1\bm{v}_1)^\top(\gamma \mathbf{e} + \bm{\eta}) + \Vert \gamma \mathbf{e} + \bm{\eta} \Vert_2^2,
\end{array}
\end{equation*}
where $\bm{A}_1 = [\bm{I} \; -\bm{I}]$ since $\bm{A}_1\bm{v}_1 = [\bm{I} \; -\bm{I}] \; \bm{v}_1 = \bm{t} - \bm{z}$. Thus, we consider the proximity term with matrix $\bm{G}_1 = c'\bm{I} - c\bm{A}_1^\top \bm{A}_1$ with $c' \geq c \cdot \lambda_{\max}(\bm{A}_1^\top\bm{A}_1)$, that is, $c' = 2Sc$. Then we have
\begin{equation*}
\begin{array}{rl}
& \displaystyle \frac{c}{2} \Vert \bm{t} - \bm{z} - \gamma \mathbf{e} - \bm{\eta} \Vert^2 + \frac{1}{2}\Vert \bm{v}_1 - \overline{\bm{v}}_1 \Vert_{\bm{G}_1} \\
= & \displaystyle \frac{c}{2} \left( \Vert \bm{A}_1\bm{v}_1 \Vert_2^2 - 2(\bm{A}_1\bm{v}_1)^\top(\gamma \mathbf{e} + \bm{\eta}) + \Vert \gamma \mathbf{e} + \bm{\eta} \Vert_2^2  \right) + \frac{c'}{2}\Vert \bm{v}_1 - \overline{\bm{v}}_1\Vert_2^2 - \frac{c}{2}(\bm{v}_1 - \overline{\bm{v}}_1)^\top \bm{A}_1^\top \bm{A}_1 (\bm{v}_1 - \overline{\bm{v}}_1) \\
= &\displaystyle \frac{c}{2} \left( - 2(\bm{A}_1\bm{v}_1)^\top(\gamma \mathbf{e} + \bm{\eta}) + \Vert \gamma \mathbf{e} + \bm{\eta} \Vert_2^2  \right)+ \frac{c'}{2}\Vert \bm{v}_1 - \overline{\bm{v}}_1\Vert_2^2 - \frac{c}{2}\left( -2 \overline{\bm{v}}_1^\top\bm{A}_1^\top \bm{A}_1\bm{v}_1 + \overline{\bm{v}}_1^\top \bm{A}_1^\top\bm{A}_1\overline{\bm{v}}_1  \right) \\
= &\displaystyle \frac{c}{2} \Vert \gamma \mathbf{e} + \bm{\eta} \Vert_2^2  - c(\bm{A}_1\bm{v}_1)^\top(\gamma \mathbf{e} + \bm{\eta}) + \frac{c'}{2}\Vert \bm{v}_1 - \overline{\bm{v}}_1\Vert_2^2 + c \cdot \overline{\bm{v}}_1^\top\bm{A}_1^\top \bm{A}_1\bm{v}_1 - \frac{c}{2}\overline{\bm{v}}_1^\top \bm{A}_1^\top\bm{A}_1\overline{\bm{v}}_1 \\
= &\displaystyle \frac{c}{2} \cdot \Vert \gamma \mathbf{e} + \bm{\eta} \Vert_2^2  + c\cdot (\bm{A}_1\overline{\bm{v}}_1 - \gamma \mathbf{e} - \bm{\eta} )^\top \bm{A}_1\bm{v}_1+ \frac{c'}{2}\Vert \bm{v}_1 - \overline{\bm{v}}_1\Vert_2^2- \frac{c}{2}\overline{\bm{v}}_1^\top \bm{A}_1^\top\bm{A}_1\overline{\bm{v}}_1,
\end{array}
\end{equation*}
where $\overline{\bm{v}}_1$ is the previous iteration value of $\bm{v}_1$, that is $\overline{\bm{v}}_1^\top = [\overline{\bm{t}}^\top \; \overline{\bm{z}}^\top ]$. Thus, the associated objective function with the above proximity term becomes
\begin{multline*}
\hat{\bm{q}}^\top\bm{t} + \theta \cdot \Vert \bm{t} \Vert_{p_*} + \bm{\sigma}^\top (\bm{t} - \bm{z}) + \sum_{s\in\mathcal{S}} \bm{\psi}_s^\top \bm{w} - \bm{\zeta}^\top \bm{z} + \frac{c}{2} \sum_{s\in\mathcal{S}} \Vert \bm{w}-\hat{\bm{w}}_s \Vert^2 \\
+ \frac{c}{2} \Vert \bm{z} -\bm{\pi}\Vert_2^2   + c\cdot (\bm{A}_1\overline{\bm{v}}_1 - \gamma \mathbf{e} - \bm{\eta} )^\top \bm{A}_1\bm{v}_1+ \frac{c'}{2}\Vert \bm{v}_1 - \overline{\bm{v}}_1\Vert_2^2.
\end{multline*}
By apply the definition of $\bm{A}_1$ and $\bm{v}_1$, we have
\begin{multline*}
\hat{\bm{q}}^\top\bm{t} + \theta \cdot \Vert \bm{t} \Vert_{p_*}  + \bm{\sigma}^\top (\bm{t} - \bm{z}) + \sum_{s\in\mathcal{S}} \bm{\psi}_s^\top \bm{w} - \bm{\zeta}^\top \bm{z} + \frac{c}{2} \sum_{s\in\mathcal{S}} \Vert \bm{w}-\hat{\bm{w}}_s \Vert_2^2 + \frac{c}{2} \Vert \bm{z} -\bm{\pi}\Vert_2^2 \\ + c\cdot (\overline{\bm{t}} - \overline{\bm{z}} - \gamma \mathbf{e} - \bm{\eta} )^\top ( \bm{t} - \bm{z}) + \frac{c'}{2}\Vert \bm{t} - \overline{\bm{t}}\Vert_2^2 + \frac{c'}{2}\Vert \bm{z} - \overline{\bm{z}}\Vert_2^2.
\end{multline*}

\subsection{Update for \texorpdfstring{$\bm{w}$}{}, \texorpdfstring{$\bm{t}$}{} and \texorpdfstring{$\bm{z}$}{}}
The above objective is seperatable in terms of $\bm{w}$, $\bm{t}$, and $\bm{z}$. We have
\begin{equation*}
\begin{array}{rl}
\mathfrak{P}_{\bm{w}}(\{\hat{\bm{w}}_s\},\{\bm{\psi}_s\})  \;\; = &\displaystyle \underset{\bm{w}}{\text{argmin}}\;  \sum_{s\in\mathcal{S}} \bm{\psi}_s^\top \bm{w} + \frac{c}{2} \sum_{s\in\mathcal{S}} \Vert \bm{w}-\hat{\bm{w}}_s \Vert^2 \\
\;\; = &\displaystyle \underset{\bm{w}}{\text{argmin}}\;  \left(\sum_{s\in\mathcal{S}} \bm{\psi}_s\right)^\top \bm{w} + \frac{c}{2} \sum_{s\in\mathcal{S}}  \bm{w}^\top \bm{w} - 2 \hat{\bm{w}}_s^\top\bm{w} \\
\;\; = &\displaystyle \underset{\bm{w}}{\text{argmin}}\; \frac{cS}{2} \bm{w}^\top \bm{w} + \left(\sum_{s\in\mathcal{S}} (\bm{\psi}_s - c \hat{\bm{w}}_s) \right)^\top \bm{w} \\
\;\; = &\displaystyle \frac{1}{cS} \left(\sum_{s\in\mathcal{S}} (c \hat{\bm{w}}_s - \bm{\psi}_s ) \right) ,
\end{array}
\end{equation*}
and
\begin{equation*}
\begin{array}{rl}
\;\;  &\displaystyle \displaystyle \mathfrak{P}_{\bm{z}}(\bm{\eta},\gamma,\{\lambda_s\},\{\bm{\alpha}_s\},\bm{\zeta},\bm{\sigma}) \\
\;\; = &\displaystyle \underset{\bm{z}}{\text{argmin}}\; \displaystyle \bm{\sigma}^\top (\bm{t} - \bm{z}) - \bm{\zeta}^\top \bm{z} + \frac{c}{2} \Vert \bm{z} -\bm{\pi}\Vert_2^2 + c\cdot (\overline{\bm{t}} - \overline{\bm{z}} - \gamma \mathbf{e} - \bm{\eta} )^\top ( \bm{t} - \bm{z})+ \frac{c'}{2}\Vert \bm{z} - \overline{\bm{z}}\Vert_2^2 \\
\;\; = & \displaystyle \underset{\bm{z}}{\text{argmin}}\;  (- \bm{\sigma} - \bm{\zeta} - c\cdot (\overline{\bm{t}} - \overline{\bm{z}} - \gamma \mathbf{e} - \bm{\eta} ) )^\top\bm{z}  + \frac{c}{2} \Vert \bm{z} -\bm{\pi}\Vert_2^2 + \frac{c'}{2}\Vert \bm{z} - \overline{\bm{z}}\Vert_2^2\\
\;\; = & \displaystyle \;\; \frac{c}{c+c'}\cdot (\bm{\pi}+\overline{\bm{t}} - \overline{\bm{z}} - \gamma \cdot \mathbf{e} - \bm{\eta}) +\frac{c'}{c+c'}\cdot \overline{\bm{z}} + \frac{1}{c+c'}\cdot(\bm{\zeta} + \bm{\sigma}) \\
\;\; = & \displaystyle \frac{1}{1+2S}\cdot \Big [ \bm{\pi}+\overline{\bm{t}} - \overline{\bm{z}} - \gamma \cdot \mathbf{e} - \bm{\eta} + 2S \cdot \overline{\bm{z}} + \frac{\bm{\zeta} + \bm{\sigma}}{c} \Big ], 
\end{array}
\end{equation*}
where the third inequality holds because the associated optimization is separable for $z_s$. By the first order condition, we have
\begin{multline*}
-\sigma_s -\zeta_s - c\cdot(\overline{t}_s - \overline{z}_s - \gamma - \eta_s) + c\cdot (z_s - \pi_s) + c' (z_s - \overline{z}_s) = 0 \\ \quad\Longleftrightarrow\quad  z_s = \frac{c\cdot \pi_s +c'\cdot \overline{z}_s + \zeta_s +\sigma_s + c\cdot(\overline{t}_s - \overline{z}_s - \gamma - \eta_s) }{c+c'}.
\end{multline*}
And the last equality holds by taking the simple substitution of $c'$. For the update of $\bm{t}$, we have 
\begin{equation*}
\begin{array}{rl}
& \displaystyle \mathfrak{P}_{\bm{t}}(\bm{\eta},\gamma,\bm{\sigma}) \\
= & \displaystyle \underset{\bm{t}}{\text{argmin}}\;  \hat{\bm{q}}^\top\bm{t} + \theta \cdot \Vert \bm{t} \Vert_{p_*} + \bm{\sigma}^\top (\bm{t} - \bm{z}) + c\cdot (\overline{\bm{t}} - \overline{\bm{z}} - \gamma \mathbf{e} - \bm{\eta} )^\top ( \bm{t} - \bm{z}) + \frac{c'}{2}\Vert \bm{t} - \overline{\bm{t}}\Vert_2^2 
\\
= & \displaystyle \underset{\bm{t}}{\text{argmin}}\; (\hat{\bm{q}} + \bm{\sigma} + c\cdot (\overline{\bm{t}} - \overline{\bm{z}} - \gamma \mathbf{e} - \bm{\eta} ))^\top\bm{t}  + \theta \cdot \Vert \bm{t} \Vert_{p_*} + \frac{c'}{2}\Vert \bm{t} - \overline{\bm{t}}\Vert_2^2 
\\
= & \displaystyle \text{prox}_{(\theta / c')\; \Vert \cdot \Vert_{p_*}} \left( \overline{\bm{t}} - \frac{\hat{\bm{q}} + \bm{\sigma} + c\cdot (\overline{\bm{t}} - \overline{\bm{z}} - \gamma \mathbf{e} - \bm{\eta})}{c'} \right)
\\
= & \displaystyle \overline{\bm{t}} - \frac{\hat{\bm{q}} + \bm{\sigma} + c\cdot (\overline{\bm{t}} - \overline{\bm{z}} - \gamma \mathbf{e} - \bm{\eta})}{c'} - \frac{\theta}{c'} \cdot\text{Proj}_{B_{\Vert \cdot \Vert_p}[0,1]} \left( \frac{c'}{\theta} \cdot \overline{\bm{t}} - \frac{1}{\theta} ( \hat{\bm{q}} + \bm{\sigma} + c\cdot (\overline{\bm{t}} - \overline{\bm{z}} - \gamma \mathbf{e} - \bm{\eta}) ) \right) \\
= & \displaystyle \overline{\bm{t}} - \frac{\hat{\bm{q}} + \bm{\sigma} + c\cdot (\overline{\bm{t}} - \overline{\bm{z}} - \gamma \mathbf{e} - \bm{\eta})}{2Sc} - \frac{\theta}{2Sc} \cdot\text{Proj}_{B_{\Vert \cdot \Vert_p}[0,1]} \left( \frac{2Sc}{\theta} \cdot \overline{\bm{t}} - \frac{1}{\theta} ( \hat{\bm{q}} + \bm{\sigma} + c\cdot (\overline{\bm{t}} - \overline{\bm{z}} - \gamma \mathbf{e} - \bm{\eta}) ) \right) \\
= & \displaystyle \bm{u} - \frac{\theta}{2Sc} \cdot \text{Proj}_{B_{\Vert \cdot \Vert_p}[0,1]} \left( \frac{2Sc}{\theta} \cdot \bm{u} \right),
\end{array}
\end{equation*}
where $\bm{u} = \overline{\bm{t}} - ( \hat{\bm{q}} + \bm{\sigma} + c\cdot (\overline{\bm{t}} - \overline{\bm{z}} - \gamma \mathbf{e} - \bm{\eta})) / (2Sc)$. Here, $\text{prox}_{f}(\cdot)$ stands for the proximal mapping operator, and the third equality holds by applying Theorem~6.13 in~\cite{beck2017first}. The fourth and fifth equality follow from Example~6.47 in~\cite{beck2017first} and substitution of $c'$. The definition of $\text{Proj}_{B_{\Vert \cdot \Vert_p}[0,1]}(\cdot)$ varies depending on the choice of different $\ell_p$ norm in the support set $\mathcal{Q}$, and we list three cases for $p = 1,2,\infty$ from Table~6.1 in~\cite{beck2017first} in the followings.
\begin{enumerate}
\item If $p = 1$, the operator of projection onto the $L_1$ ball is defined as
\begin{equation*}
\text{Proj}_{B_{\Vert \cdot \Vert_1}[0,1]} (\bm{x}) = \left\{ 
\begin{array}{ll}
\displaystyle \bm{x}, & \text{if} \Vert \bm{x} \Vert_1 \leq 1 \\
\displaystyle \mathcal{T}_{\delta^*}(\bm{x}) & \text{if} \Vert \bm{x} \Vert_1 > 1,
\end{array}
\right.
\end{equation*}
where $\delta^*$ is any positive root of the nonincreasing function $\varphi(\delta) = \Vert \mathcal{T}_{\delta}(\bm{x}) \Vert_1 - 1$. $\mathcal{T}_{\delta}$ is the soft thresholding operator given by $\mathcal{T}_{\delta}(\bm{x}) = [\bm{x}-\delta\bm{e}]_+ \odot \text{sgn}(\bm{x})$, where $a \odot b$ represents Hadamard product and $\text{sgn}(\bm{x})$ denotes the sign vector of $\bm{x}$.
\item If $p = 2$, the operator of projection onto the $L_2$ ball is defined as
\begin{equation*}
\text{Proj}_{B_{\Vert \cdot \Vert_2}[0,1]} (\bm{x}) = \frac{1}{\max\{ \Vert \bm{x} \Vert_2, 1 \}} \cdot \bm{x}.
\end{equation*}
\item If $p = \infty$, we transform set $B_{\Vert \cdot \Vert_\infty}[0,1]$ into the equivalent set $\text{Box}[-\mathbf{e}, \mathbf{e}]$. According to the definition of the projection onto the $\text{Box}[-\mathbf{e}, \mathbf{e}]$, we define
\begin{equation*}
\text{Proj}_{B_{\Vert \cdot \Vert_\infty}[0,1]} (\bm{x}) = \text{Proj}_{\text{Box}[-\mathbf{e}, \mathbf{e}]} (\bm{x}) = \min \{ \max \{ \bm{x}, -\mathbf{e} \}, \mathbf{e} \}.
\end{equation*}
\end{enumerate}

\subsection{Adding Proximal Term for Updating \texorpdfstring{$(\bm{\eta},\gamma,\{\lambda_s\},\{\bm{\alpha}_s\},\{\hat{\bm{w}}_s\})$}{}}
The optimization problem associated with the update for the second group $(\bm{\eta},\gamma,\{\lambda_s\},\{\bm{\alpha}_s\},\{\hat{\bm{w}}_s\})$ has the objective function
\begin{multline*}
- \gamma - \sum_{s\in\mathcal{S}} \bm{\psi}_s^\top \hat{\bm{w}}_s  + \sum_{s\in\mathcal{S}} \zeta_s \left( \rho_s \lambda_s + \frac{1}{N_s} \mathbf{e}^\top \bm{\alpha}_{s} \right) + \bm{\sigma}^\top ( - \gamma \mathbf{e} - \bm{\eta}) + \frac{c}{2} \sum_{s\in\mathcal{S}} \Vert \bm{w}-\hat{\bm{w}}_s \Vert^2 \\ + \frac{c}{2} \sum_{s\in\mathcal{S}} \left( \rho_s \lambda_s + \frac{1}{N_s} \mathbf{e}^\top \bm{\alpha}_{s} - z_s \right)^2 + \frac{c}{2} \Vert \bm{t} - \bm{z} - \gamma \mathbf{e} - \bm{\eta} \Vert^2.
\end{multline*}
We apply the same approach as updating the variables of the first group to $(\bm{\eta},\gamma)$, while directly solve a convex problem for updating $(\{\lambda_s\},\{\bm{\alpha}_s\},\{\hat{\bm{w}}_s\})$. First, we decompose into two independent sub-problems for $(\bm{\eta},\gamma)$.

As for $\bm{\eta}$ and $\gamma$, we have the objective 
\begin{equation*}
- \gamma  + \bm{\sigma}^\top ( - \gamma \mathbf{e} - \bm{\eta})  + \frac{c}{2} \Vert \bm{t} - \bm{z} - \gamma \mathbf{e} - \bm{\eta} \Vert^2.
\end{equation*}
To efficiently compute the update, we add a proximity term as in AD-PMM. In particular, we denote $\bm{v}_2^\top = [\gamma \; \bm{\eta}^\top ]$, and we have
\begin{equation*}
\begin{array}{ll}
\displaystyle \Vert \bm{t} - \bm{z} - \gamma \mathbf{e} - \bm{\eta} \Vert^2 
& \displaystyle = \Vert \bm{t} - \bm{z} \Vert_2^2 + 2(\bm{t} - \bm{z} )^\top( - \gamma \mathbf{e} - \bm{\eta}) + \Vert - \gamma \mathbf{e} - \bm{\eta} \Vert_2^2 \\
& \displaystyle = \Vert \bm{t}-\bm{z} \Vert_2^2 + 2(\bm{t}-\bm{z})^\top\bm{A}_2\bm{v}_2  + \Vert \bm{A}_2\bm{v}_2 \Vert_2^2, 
\end{array}
\end{equation*}
where $\bm{A}_2 = [-\mathbf{e} \; -\bm{I}]$ since $\bm{A}_2\bm{v}_2 = [-\mathbf{e} \; -\bm{I}] \; \bm{v}_2 = -\gamma\mathbf{e} - \bm{\eta}$. Thus, we consider the proximity term with matrix $\bm{G}_2 = c''\bm{I} - c\bm{A}_2^\top \bm{A}_2$ with $c'' \geq c \cdot \lambda_{\max}(\bm{A}_2^\top\bm{A}_2)$, that is $c'' = 2Sc$. Then we have
\begin{equation*}
\begin{array}{rl}
& \displaystyle \frac{c}{2} \Vert \bm{t} - \bm{z} - \gamma \mathbf{e} - \bm{\eta} \Vert^2 + \frac{1}{2}\Vert \bm{v}_2 - \overline{\bm{v}}_2 \Vert_{\bm{G}_2} \\
= & \displaystyle \frac{c}{2} \left( \Vert \bm{t}-\bm{z} \Vert_2^2 + 2(\bm{t}-\bm{z})^\top\bm{A}_2\bm{v}_2  + \Vert \bm{A}_2\bm{v}_2 \Vert_2^2   \right) + \frac{c''}{2}\Vert \bm{v}_2 - \overline{\bm{v}}_2\Vert_2^2 - \frac{c}{2}(\bm{v}_2 - \overline{\bm{v}}_2)^\top \bm{A}_2^\top \bm{A}_2 (\bm{v}_2 - \overline{\bm{v}}_2) 
 \\
= &\displaystyle \frac{c}{2} \left( \Vert \bm{t}-\bm{z} \Vert_2^2 + 2(\bm{t}-\bm{z})^\top\bm{A}_2\bm{v}_2 \right)+ \frac{c''}{2}\Vert \bm{v}_2 - \overline{\bm{v}}_2\Vert_2^2 - \frac{c}{2}\left( -2 \overline{\bm{v}}_2^\top\bm{A}_2^\top \bm{A}_2\bm{v}_2 + \overline{\bm{v}}_2^\top \bm{A}_2^\top\bm{A}_2\overline{\bm{v}}_2  \right) 
\\
= &\displaystyle \frac{c}{2} \cdot \Vert \bm{t}-\bm{z} \Vert_2^2  + c(\bm{t}-\bm{z})^\top\bm{A}_2\bm{v}_2 + \frac{c''}{2}\Vert \bm{v}_2 - \overline{\bm{v}}_2\Vert_2^2 + c \cdot \overline{\bm{v}}_2^\top\bm{A}_2^\top \bm{A}_2\bm{v}_2 - \frac{c}{2}\overline{\bm{v}}_2^\top \bm{A}_2^\top\bm{A}_2\overline{\bm{v}}_2 \\
= &\displaystyle \frac{c}{2} \cdot \Vert \bm{t}-\bm{z} \Vert_2^2  + c\cdot (\bm{A}_2\overline{\bm{v}}_2 + \bm{t} - \bm{z} )^\top \bm{A}_2\bm{v}_2+ \frac{c''}{2}\Vert \bm{v}_2 - \overline{\bm{v}}_2\Vert_2^2- \frac{c}{2}\overline{\bm{v}}_2^\top \bm{A}_2^\top\bm{A}_2\overline{\bm{v}}_2,
\end{array}
\end{equation*}
where $\overline{\bm{v}}_2$ is the previous iteration value of $\bm{v}_2$, that is $\overline{\bm{v}}_2^\top = [\overline{\gamma} \; \overline{\bm{\eta}}^\top ]$. Thus, the associated objective function with the above proximity term becomes
\begin{equation*}
- \gamma  + \bm{\sigma}^\top ( - \gamma \mathbf{e} - \bm{\eta})  + c\cdot (\bm{A}_2\overline{\bm{v}}_2 + \bm{t} - \bm{z} )^\top \bm{A}_2\bm{v}_2+ \frac{c''}{2}\Vert \bm{v}_2 - \overline{\bm{v}}_2\Vert_2^2.
\end{equation*}
By apply the definition of $\bm{A}_2$ and $\bm{v}_2$, we have
\begin{equation*}
- \gamma  + \bm{\sigma}^\top ( - \gamma \mathbf{e} - \bm{\eta})  + c\cdot (- \overline{\gamma}\mathbf{e} - \overline{\bm{\eta}} + \bm{t} - \bm{z} )^\top (-\gamma\mathbf{e} - \bm{\eta}) + \frac{c''}{2}\Vert \gamma - \overline{\gamma}\Vert_2^2 + \frac{c''}{2}\Vert \bm{\eta} - \overline{\bm{\eta}}\Vert_2^2.
\end{equation*}

\subsection{Update for \texorpdfstring{$\bm{\eta}$}{} and \texorpdfstring{$\gamma$}{}}
The above objective is seperatable in terms of $\bm{\eta}$ and $\gamma$, and so we have
\begin{equation*}
\begin{array}{rl}
\displaystyle \mathfrak{P}_{\bm{\eta}}(\bm{t},\bm{z},\bm{\sigma})  \;\; = &\displaystyle \underset{\bm{\eta} \in \mathbb{R}^S_+ }{\text{argmin}}\;  
 -\bm{\sigma}^\top\bm{\eta} - c\cdot (- \overline{\gamma}\mathbf{e} - \overline{\bm{\eta}} + \bm{t} - \bm{z} )^\top \bm{\eta} + \frac{c''}{2}\Vert \bm{\eta} - \overline{\bm{\eta}}\Vert_2^2  \\
\;\; = & \displaystyle \underset{\bm{\eta}\in \mathbb{R}^S_+}{\text{argmin}}\;  
(-\bm{\sigma} - c\cdot (- \overline{\gamma}\mathbf{e} - \overline{\bm{\eta}} + \bm{t} - \bm{z} ))^\top\bm{\eta} + \frac{c''}{2}\Vert \bm{\eta} - \overline{\bm{\eta}}\Vert_2^2 \\
\;\; = & \displaystyle \left[ \overline{\bm{\eta}} + \frac{1}{c''} \cdot \bm{\sigma} + \frac{c}{c''} \cdot (-\overline{\gamma} \cdot \mathbf{e} - \overline{\bm{\eta}} + \bm{t} - \bm{z}) \right]_+ \\
\;\; = & \displaystyle \left[ \overline{\bm{\eta}} + \frac{1}{2S} \cdot \left( \frac{1}{c}\cdot \bm{\sigma} -\overline{\gamma} \cdot \mathbf{e} - \overline{\bm{\eta}} + \bm{t} - \bm{z}\right) \right]_+ .
\end{array}
\end{equation*}
Here, the third inequality holds because the associated optimization is separable for $\eta_s$. By the first order condition, we have
\begin{equation*}
(-\sigma_s - c \cdot(-\overline{\gamma} - \overline{\eta}_s + t_s - z_s)) + c''(\eta_s - \overline{\eta}_s) = 0 
\quad\Longleftrightarrow\quad  
\eta_s = \frac{c''\cdot \overline{\eta}_s +\sigma_s + c \cdot(-\overline{\gamma} - \overline{\eta}_s + t_s - z_s)}{c''}.
\end{equation*}
And the last equality holds by taking the substitution of $c''$. Similarly by applying the first order condition, for $\gamma$, we have
\begin{equation*}
\begin{array}{rl}
\displaystyle \mathfrak{P}_{\gamma}(\bm{t},\bm{z},\bm{\sigma})  \;\; = &\displaystyle \underset{\gamma}{\text{argmin}}\;  
 - \gamma  - \gamma \cdot \bm{\sigma}^\top \mathbf{e} - c\gamma\cdot (- \overline{\gamma}\mathbf{e} - \overline{\bm{\eta}} + \bm{t} - \bm{z} )^\top\mathbf{e} + \frac{c''}{2}\Vert \gamma - \overline{\gamma}\Vert_2^2  \\
\;\; = & \displaystyle \underset{\gamma}{\text{argmin}}\;  
-\gamma \cdot (1 + \bm{\sigma}^\top \mathbf{e} + c\cdot (- \overline{\gamma}\mathbf{e} - \overline{\bm{\eta}} + \bm{t} - \bm{z} )^\top\mathbf{e} )  + \frac{c''}{2}\Vert \gamma - \overline{\gamma}\Vert_2^2
\\
\;\; = & \displaystyle \overline{\gamma} + \frac{1}{c''} \cdot \Big [ 1 + \bm{\sigma}^\top \mathbf{e} + c\cdot (- \overline{\gamma}\mathbf{e} - \overline{\bm{\eta}} + \bm{t} - \bm{z} )^\top\mathbf{e} \Big ] \\
\;\; = & \displaystyle \overline{\gamma} + \frac{1}{2Sc} \cdot \Big [ 1 + \bm{\sigma}^\top \mathbf{e} + c\cdot (- \overline{\gamma}\mathbf{e} - \overline{\bm{\eta}} + \bm{t} - \bm{z} )^\top\mathbf{e} \Big ],
\end{array}
\end{equation*}
where the last equality follows from the substitution of $c''$.

\subsection{Update for \texorpdfstring{$\{\lambda_s\}$}{}, \texorpdfstring{$\{\bm{\alpha}_s\}$}{} and \texorpdfstring{$\{\hat{\bm{w}}_s\}$}{}}
In terms of $(\{\lambda_s\},\{\bm{\alpha}_s\},\{\hat{\bm{w}}_s\})$, we have the optimization problem
\begin{multline*}
\underset{ (\lambda_s,\bm{\alpha}_s,\hat{\bm{w}}_s)\in \Omega_s,\; s\in\mathcal{S} }{\text{min}}\;  - \sum_{s\in\mathcal{S}} \bm{\psi}_s^\top \hat{\bm{w}}_s + \frac{c}{2} \sum_{s\in\mathcal{S}} \Vert \bm{w}-\hat{\bm{w}}_s \Vert^2 + \sum_{s\in\mathcal{S}} \zeta_s \left( \rho_s \lambda_s + \frac{1}{N_s} \mathbf{e}^\top \bm{\alpha}_{s}  \right) \\ + \frac{c}{2} \sum_{s\in\mathcal{S}} \left( \rho_s \lambda_s + \frac{1}{N_s} \mathbf{e}^\top \bm{\alpha}_{s} - z_s \right)^2.
\end{multline*}
Obviously, the above problem is separable for $(\lambda_s,\bm{\alpha}_s,\hat{\bm{w}}_s)$, which enables us to solve the convex optimization sub-problem $\mathfrak{C}_s (\bm{w}_s,z_s,\bm{\psi}_s,\zeta_s)$ shown in the following, using existing solvers for each $s\in\mathcal{S}$,
\begin{equation*}
\underset{(\lambda_s,\bm{\alpha}_s,\hat{\bm{w}}_s)\in \Omega_s}{\text{argmin}}\; -  \bm{\psi}_s^\top \hat{\bm{w}}_s + \frac{c}{2} \cdot \Vert \bm{w}-\hat{\bm{w}}_s \Vert^2 +  \zeta_s \left( \rho_s \lambda_s + \frac{1}{N_s} \mathbf{e}^\top \bm{\alpha}_{s}  \right) + \frac{c}{2}  \left( \rho_s \lambda_s + \frac{1}{N_s} \mathbf{e}^\top \bm{\alpha}_{s} - z_s \right)^2.
\end{equation*}

\end{document}